\documentclass[10pt,twocolumn,letterpaper]{article}
\usepackage[]{cvpr}

\usepackage{url}
\usepackage{graphicx}
\usepackage{tabularx}       
\usepackage{comment}
\usepackage{makecell}
\usepackage{pifont}
\usepackage{booktabs}
\usepackage{multirow}

\usepackage[misc]{ifsym}
\usepackage{afterpage}
\usepackage{setspace}
\usepackage{breakcites}
\usepackage{seqsplit}

\usepackage{times}
\usepackage{epsfig}
\usepackage{graphicx}
\usepackage{amsmath}
\usepackage{amssymb}
\usepackage{float}
\usepackage[symbol]{footmisc}

\usepackage{hyphenat}
\usepackage{soul,color}
\usepackage{amsopn}

\definecolor{Red}{cmyk}{0,1,1,0}
\definecolor{Green}{cmyk}{1,0,1,0}
\definecolor{Cyan}{cmyk}{1,0,0,0}
\definecolor{Purple}{cmyk}{0.45,0.86,0,0}
\definecolor{Rosolic}{cmyk}{0.00,1.00,0.50,0}
\definecolor{Blue}{cmyk}{1.00,1.00,0.00,0}
\definecolor{BlueViolet}{cmyk}{0.86,0.91,0,0.04}
\definecolor{NavyBlue}{cmyk}{0.94,0.54,0,0}

\newcommand{\cx}[1]{{#1}}

\newcommand{\hidden}[1]{{\color{NavyBlue}}}

\newcommand{\jb}[1]{{#1}}

\newcommand{\myparagraph}[1]{\vspace{0.1em}\noindent\textbf{#1}}

\usepackage[pagebackref=true,breaklinks=true,colorlinks,bookmarks=false]{hyperref}

\usepackage[capitalize]{cleveref}
\crefname{section}{Sec.}{Secs.}
\Crefname{section}{Section}{Sections}
\Crefname{table}{Table}{Tables}
\crefname{table}{Tab.}{Tabs.}

\def\cvprPaperID{150} 
\def\confName{CVPR}
\def\confYear{CVPR}

\begin{document}

\title{Executing your Commands via Motion Diffusion in Latent Space}


\author{Xin Chen$^{1}$\footnotemark[1]\quad\quad\quad\quad Biao Jiang$^{2}$\footnotemark[1] \quad\quad\quad\quad Wen Liu$^{1}$ \quad\quad\quad\quad Zilong Huang$^{1}$  \quad\quad\quad\quad Bin Fu$^{1}$ \\ Tao Chen$^{2}$ \stepcounter{footnote} 
\quad\quad\quad\quad Gang Yu$^{1}$\footnotemark[2]
    \\ $^{1}$Tencent PCG \quad\quad\quad\quad $^{2}$Fudan University\\ 
\tt \small \textbf{\href{https://github.com/chenfengye/motion-latent-diffusion}{https://github.com/chenfengye/motion-latent-diffusion}}}


\maketitle

\footnotetext[1]{These authors contributed equally to this work.}
\footnotetext[2]{Corresponding author.}

\begin{abstract}
    \vspace{-5pt}
    We study a challenging task, conditional human motion generation, which produces plausible human motion sequences according to various conditional inputs, such as action classes or textual descriptors. Since human motions are highly diverse and have a property of quite different distribution from conditional modalities, such as textual descriptors in natural languages, it is hard to learn a probabilistic mapping from the desired conditional modality to the human motion sequences. Besides, the raw motion data from the motion capture system might be redundant in sequences and contain noises; directly modeling the joint distribution over the raw motion sequences and conditional modalities would need a heavy computational overhead and might result in artifacts introduced by the captured noises. To learn a better representation of the various human motion sequences, we first design a powerful Variational AutoEncoder (VAE) and arrive at a representative and low-dimensional latent code for a human motion sequence. Then, instead of using a diffusion model to establish the connections between the raw motion sequences and the conditional inputs, we perform a diffusion process on the motion latent space. Our proposed Motion Latent-based Diffusion model (MLD) could produce vivid motion sequences conforming to the given conditional inputs and substantially reduce the computational overhead in both the training and inference stages. Extensive experiments on various human motion generation tasks demonstrate that our MLD achieves significant improvements over the state-of-the-art methods among extensive human motion generation tasks, with two orders of magnitude faster than previous diffusion models on raw motion sequences.


\end{abstract}


\begin{figure}[tbp] 
	\centering 
	\includegraphics[width=0.9\linewidth]{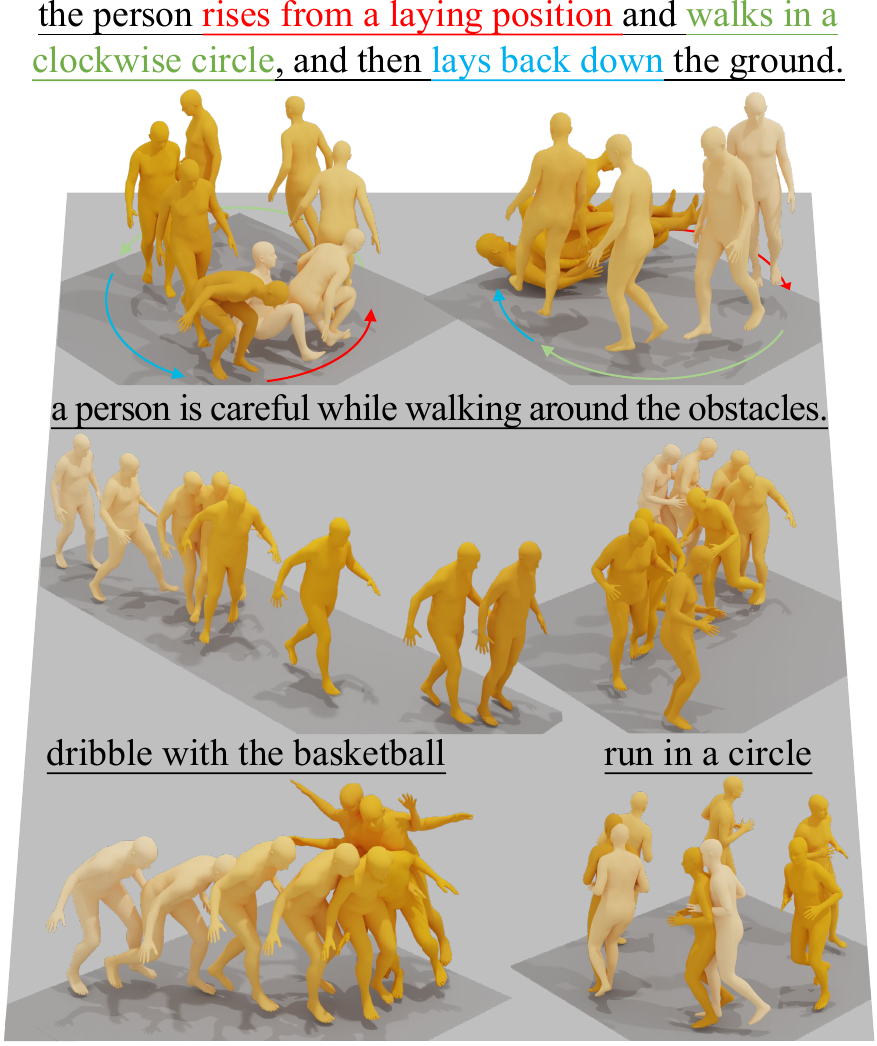} 
	\caption{Our Motion Latent-based Diffusion (MLD) model can achieve high-quality and diverse motion generation given a text prompt. The darker colors indicate the later in time, and the colored words refer to the motions with same colored trajectory.} 
	\label{fig:teaser} 
\end{figure} 

\section{Introduction}
\label{intro}
Human motion synthesis has recently rapidly developed in a multi-modal generative fashion. Various condition inputs, such as music~\cite{li2021ai, li2022danceformer, lee2019dancing}, control signals~\cite{2021-TOG-AMP, starke2019neural, starke2022deepphase}, action categories~\cite{petrovich21actor, guo2020action2motion}, and natural language descriptions~\cite{Guo_2022_CVPR_t2m,petrovich22temos, mdm2022human,chuan2022tm2t,ahuja2019language2pose,kim2022flame}, provide a more convenient and human-friendly way to animate virtual characters or even control humanoid robots. It will benefit numerous applications in the game industry, film production, VR/AR, and robotic assistance.

Among all conditional modalities, text-based conditional human motion synthesis has been driving and dominating research frontiers because the language descriptors provide a convenient and natural user interface for people to interact with computers~\cite{petrovich22temos, ahuja2019language2pose, zhang2022motiondiffuse, mdm2022human, kim2022flame}. However, since the distributions between the natural language descriptors and motion sequences are quite different, it is not easy to learn a probabilistic mapping function from the textural descriptors to the motion sequences, which is also mentioned in the previous work, MotionCLIP~\cite{tevet2022motionclip}. Two typical methods address this problem: 1) the cross-modal compatible latent space between motion and language~\cite{petrovich22temos, ahuja2019language2pose} and 2) the conditional diffusion model~\cite{zhang2022motiondiffuse, mdm2022human, kim2022flame}. The formers, such as TEMOS~\cite{petrovich22temos}, usually learn a motion Variational AutoEncoder (VAE) and a text Variational Encoding (without decoder) and then constrain the text encoder and the motion encoder into a compatible latent space via the Kullback-Leibler (KL) divergences loss, which pushes a foundational step forward on creating human motion sequences by natural language inputs. However, since the distributions of natural languages and motion sequences are highly different, forcibly aligning these two simple gaussian distributions, in terms of variational text encoding and variational motion encoding, into a compatible distribution might result in misalignments and thereby reduce the generative diversity inevitably. In light of the tremendous success of the diffusion-based generative models on other domains~\cite{ramesh2022hierarchical,imagen_saharia2022photorealistic, stable_diffusion,ho2022imagen, Zhou_2021_ICCV, xu2022geodiff}, the latter category methods~\cite{zhang2022motiondiffuse, mdm2022human, kim2022flame} propose a conditional diffusion model for human motion synthesis to learn a more powerful probabilistic mapping from the textual descriptors to human motion sequences and improve the synthesized quality and diversity. Nevertheless, the raw motion sequences are somewhat time-axis redundant, and diffusion models in raw sequential data~\cite{rasul2021autoregressive, ho2022imagen, li2022diffusion} usually require exhausting computational overhead in both the training and inference phase, which is inefficient. Besides, since the raw motion data from the motion capture system might contain noises, the powerful diffusion models might learn the clues of a probabilistic mapping from the conditional inputs to the noise motion sequences and produce artifacts.

To efficiently synthesize plausible and diverse human motion sequences according to the conditional inputs, inspired by the success of the diffusion model on latent space in text-to-image synthesis~\cite{stable_diffusion}, we combine the advantages of the latent space-based and the conditional diffusion-based methods and propose a motion latent-based diffusion model (MLD) for human motion generation. Specifically, we first design a transformer-based autoencoder~\cite{petrovich21actor} with the UNet-like long skip connections~\cite{ronneberger2015u} to learn a representative and low-dimensional latent distribution of human motion sequences. Then, instead of using a diffusion model to establish the connections between the raw motion sequences and the conditional inputs, we propose a motion latent-based diffusion model (MLD) to learn a better probabilistic mapping from the conditions to the representative motion latent codes, which could not only produce the vivid motion sequences conforming to the given conditional inputs but also substantially reduce the computational overhead in both training and inference stage. In addition, high-quality human motion sequences with well-annotated action labels or textual descriptions are expensive and limited. In contrast, the large-scale non-annotated or weakly annotated motion sequences are publicly available, such as the AMASS dataset~\cite{AMASS_ICCV2019}. Our proposed MLD could individually train a motion latent autoencoder on these large-scale datasets, arriving at a representative and low-dimensional latent space for diverse human motion sequences. This low-dimensional latent space with higher information density could accelerate the model's convergence and significantly reduce computational consumption for the downstream conditional human motion generation tasks.

We summarize the contributions as follows: 1) we design and explore a more representative motion variational autoencoder (VAE), which provides state-of-the-art motion reconstruction and diverse generation, benefiting the training of the latent diffusion models; 2) we further demonstrate that motion generation tasks on latent spaces, such as text-to-motion and action-to-motion, are more efficient than the diffusion models on raw motion sequences; 3) our proposed MLD achieves competitive performance on multiple tasks (unconditional motion generation, action-to-motion, and text-to-motion), and codes are available.

\section{Related Work}
\label{relatedwork}

\textbf{Human Motion Synthesis} allows rich inputs of multi-modal data, such as text~\cite{Guo_2022_CVPR_t2m,petrovich22temos, mdm2022human,chuan2022tm2t,ahuja2019language2pose,kim2022flame}, action category~\cite{petrovich21actor, guo2020action2motion}, incomplete pose sequences~\cite{duan2021single, harvey2020robust, mdm2022human}, control signals~\cite{starke2022deepphase, starke2019neural,2021-TOG-AMP}, musics~\cite{li2021ai, li2022danceformer, lee2019dancing} and image(s)~\cite{rempe2021humor, chen2022learning},
here, we focus on some typical tasks.
Firstly, \textbf{unconditional motion generation}~\cite{yan2019convolutional,zhao2020bayesian,zhang2020perpetual, raab2022modi, mdm2022human} is a more universal task, which models the entire motion space, only needs motion data without any requirement of annotation, and benefits other generation tasks.
VPoser~\cite{vposer_SMPL-X:2019} proposes a variational human pose prior mainly for image-based pose fitting.
ACTOR~\cite{petrovich21actor, petrovich22temos} recently proposes a class-agnostic transformer VAE as one baseline.
After that, among all conditional tasks, \textbf{text-to-motion} \cite{petrovich22temos, ahuja2019language2pose, zhang2022motiondiffuse, mdm2022human, kim2022flame, Guo_2022_CVPR_t2m} has been driving and dominating
research frontiers because the language descriptors are the most user-friendly and convenient.
More recently, two categories of motion synthesis methods have emerged, joint-latent models~\cite{petrovich22temos, ahuja2019language2pose} and diffusion models~\cite{zhang2022motiondiffuse, mdm2022human, kim2022flame}.
The former category, like TEMOS~\cite{petrovich22temos}, proposes a VAE architecture to learn a joint latent space of motion and text constrained on a Gaussian distribution.
However, natural language and human motions are quite different with misaligned structure and distribution, thus it is difficult to forcibly align two simple Gaussian distributions ~\cite{tevet2022motionclip}.
Lastly, we introduce \textbf{action-to-motion}~\cite{petrovich21actor, guo2020action2motion}, a reverse problem of the classical action recognition task.
ACTOR~\cite{petrovich21actor} proposes learnable biases in transformer VAE to embed action for motion generation.
However, most above methods can only handle one task and hardly change condition inputs.
%
We address this problem by separating models into a universal motion generative model and latent diffusion models to handle different motion generation tasks.


\textbf{Motion data} is critical in the development of motion synthesis tasks.
Thanks to the marker-based and markless motion capture approaches~\cite{VIBE_CVPR2020, wan2021encoder, he2021challencap, chen2021sportscap}, they provide convenient and effective solutions for large raw motion data collection.
KIT Motion-Language~\cite{Plappert2016kit} annotates sequence-level description for motions from ~\cite{mandery2015kit}, and HumanML3d~\cite{Guo_2022_CVPR_humanml3d} provide more textual annotation for some motions of AMASS\cite{AMASS_ICCV2019}.
They are also our focus in the text-to-motion task.
For the action-to-motion datasets, Babel~\cite{BABEL:CVPR:2021} also collects motions from AMASS and provides action and behavior annotations.
ACTOR~\cite{petrovich21actor} use \cite{VIBE_CVPR2020} to process two action recognition datasets, HumanAct12~\cite{guo2020action2motion} and UESTC~\cite{ji2018large}, for action-to-motion task.

\textbf{Motion Representation.} These datasets lead to the discussion about motion representation, such as the straightforward joint positions and the Master Motor Map (MMM) format~\cite{terlemez2014master}.
For our setting, we employ two motion representations: 1) the classical SMPL-based~\cite{SMPL2015, VIBE_CVPR2020, chen2021sportscap} motion parameters and 2) the redundant hand-crafted motion feature~\cite{Guo_2022_CVPR_humanml3d,starke2019neural,starke2022deepphase} with a combination of joints features.
The former is widely used in motion capture, and the latter is mainly used in character animation.
As suggested by \cite{Guo_2022_CVPR_humanml3d}, we use the latter in most of our synthesis framework to avoid foot-sliding issues, and use the SMPL parameters for the action-based tasks for a fair comparison with other approaches.
Besides, we also recognize the latent in \cref{sec:exp:vae} as one of motion representation.

\textbf{Generative Models} play an important role in motion synthesis tasks to generate high-quality human motion, 
Although motion generative models, like VAEs~\cite{petrovich21actor,motionvae_ling2020character, guo2020action2motion} and Generative Adversarial Networks (GAN)~\cite{lin2018human, ahn2018text2action}, can enable effective human motion sampling,
%
%
recent studies~\cite{arjovsky2017wasserstein, gulrajani2017improved, guo2020action2motion, petrovich21actor} recommend VAEs rather than GANs since the latter are more difficult to train.
We follow their suggestions and employ VAEs to compress and reconstruct human motion for the learning of diffusion models.
We next introduce the diffusion models, especially in motion domain.

\textbf{Diffusion Generative Models.}
Diffusion Generative Models~\cite{sohl2015deep} achieve significant success in the image synthesis domain, such as Imagen~\cite{imagen_saharia2022photorealistic}, DALL·E 2~\cite{ramesh2022hierarchical} and Stable Diffusion~\cite{stable_diffusion}.
Inspired by their works, most recent methods~\cite{mdm2022human, zhang2022motiondiffuse, kim2022flame} leverage diffusion models for human motion synthesis.
MotionDiffuse~\cite{zhang2022motiondiffuse} is the first text-based motion diffusion model with fine-grained instructions on body parts.
MDM~\cite{mdm2022human}, most recently, proposes a motion diffusion model on raw motion data to learn the relation between motion and input conditions.
However, these diffusion models are not very applicable to raw motion data with potential noise and temporal consistency redundancy and thus are easily misdirected by outliers.
In addition, directly applying the diffusion model~\cite{mdm2022human, zhang2022motiondiffuse} to the raw motion data suffers from high computational overheads and low inference speed. Inspired by~\cite{stable_diffusion}, we propose a motion latent-based diffusion model to reduce computational resources and improve the generative quality.

\section{Overview}
Our goal is to capture challenging 3D human motions from a single RGB video, which suffers from extreme poses, complex motion patterns and severe self-occlusion.
Fig.~\ref{fig:overview} provides an overview of ChallenCap, which relies on a template mesh of the actor and makes full usage of multi-modal references in a learning-and-optimization framework.
Our method consists of a hybrid motion inference module to learn the challenging motion characteristics from the supervised references modalities, and a robust motion optimization module to further extract the reliable motion hints in the input images for more accurate tracking.

\myparagraph{Template and Motion Representation.}
We use a 3D body scanner to generate the template mesh of the actor and rig it by fitting the Skinned Multi-Person Linear Model (SMPL)\cite{SMPL2015} to the template mesh and transferring the SMPL skinning weights to our scanned mesh.
The kinematic skeleton is parameterized as $\textbf{S}=[\boldsymbol{\theta}, \textbf{R},\textbf{t}]$, including the joint angles $\boldsymbol{\theta} \in \mathbb{R}^{30}$ of the $N_J$ joints, the global rotation $ \textbf{R}\in\mathbb{R}^3$ and translation $\textbf{t} \in \mathbb{R}^{3}$ of the root.
Furthermore, let $\textbf{Q}$ denotes the quaternions representation of the skeleton.
Thus, we can formulate $\textbf{S}=\mathcal{M}(\textbf{Q},\textbf{t})$ where $\mathcal{M}$ denotes the motion transformation between various representations.


\myparagraph{Hybrid Motion Inference.}
Our novel motion inference scheme extracts the challenging motion characteristics from the supervised marker-based and sparse multi-view references in a data-driven manner.
We first obtain the initial noisy skeletal motion map from the monocular video.
Then, a novel generation network, HybridNet, is adopted to boost the initial motion map, which consists of a temporal encoder-decoder to extract local and global motion details from the sparse-view references, as well as a motion discriminator to utilize the motion characteristics from the unpaired marker-based references.
To train our HybridNet, a new dataset with rich references modalities and various challenging motions is introduced (Sec. \ref{sec:mot_inference}).


\myparagraph{Robust Motion Optimization.}
Besides the data-driven 3D motion characteristics from the previous stage, the input RGB video also encodes
reliable motion hints for those non-extreme poses, especially for the non-occluded regions.
Thus, a robust motion optimization is introduced to refine the skeletal motions so as to increase the tracking accuracy and overlay performance, which jointly utilizes the learned 3D prior from the supervised multi-modal references as well as the reliable 2D and silhouette information from the input image references (Sec. \ref{sec:mot_optimization}).




\section{Method}
\label{method}

To efficiently generate high-quality and diverse human motion sequences according to desired conditional inputs with fewer computational overheads, we propose to perform a diffusion process on a representative and low-dimensional motion latent space and consequently arrive at a motion latent-based diffusion model (MLD) for conditional human motion synthesis. It contains a motion Variational AutoEncoder (VAE) to learn a representative and low-dimensional latent space for diverse human motion sequences (details in \cref{sec:method:vae}) and a conditional diffusion model in this latent space (details in  \cref{sec:method:diffusion} and ~\cref{sec:method:condition}).


The conditions include action labels, textual descriptions, or even empty conditions. Specifically, given an input condition $c$, such as a sentence $\boldsymbol {w}^{1:N}=\{w^i\}_{i=1}^{N}$ describing a motion ~\cite{petrovich22temos},
a action label $a$ from the predefined action categories set $a \in A$~\cite{petrovich21actor} or even a empty condition $c = \varnothing$~\cite{vposer_SMPL-X:2019, zhang2021we},
our MLD aims to generate a human motion $\hat{x}^{1:L}=\{\hat{x}^i\}_{i=1}^{L}$ in a non-deterministic way, where L denotes the motion length or frame number.
Here, we employ the motion representation in \cite{Guo_2022_CVPR_humanml3d}: a combination of 3D joint rotations, positions, velocities, and foot contact.
In addition, we propose the motion encoder $\mathcal{E}$ to encode the motion sequences, ${x}^{1:L}=\{{x}^i\}_{i=1}^{L}$, into a latent $z = \mathcal{E}(x^{1:L})$, and decode $z$ into the motion sequences using a motion decoder $\mathcal{D}$, that is $\hat{x}^{1:L} = \mathcal{D}(z) = \mathcal{D}(\mathcal{E}(x^{1:L}))$.



\subsection{Motion Representation in Latents}
\label{sec:method:vae}

We build our motion Variational AutoEncoder, $\mathcal{V}$, based on a transformer-based architecture~\cite{petrovich21actor}, which consists of a transformer encoder $\mathcal{E}$ and a transformer decoder $\mathcal{D}$. The motion VAE, $\mathcal{V}=\{\mathcal{E}, \mathcal{D}\}$, is trained by the motion $x^{1:L}$ reconstruction only with the Mean Squared Error (MSE) loss and the Kullback-Leibler (KL) loss. We further enhance two transformers~\cite{vaswani2017attention} of $\mathcal{E}$ and $\mathcal{D}$ with long skip connections~\cite{ronneberger2015u}, and remove the action biases used in \cite{petrovich21actor}. The encoder could produce a representative, low-dimensional latent space with high informative density, and the decoder could well reconstruct the latent into motion sequences.

More specifically, the motion encoder $\mathcal{E}$ takes as input learnable distribution tokens, and frame-wise motion features $x^{1:L}$ of arbitrary length $L$. We use the embedded distribution tokens as Gaussian distribution parameters $\mu$ and $\sigma$ of the motion latent space $\mathcal{Z}$ to reparameterize~\cite{kingma2013auto} latent $z \in \mathbb{R}^{ n\times d}$ whose dimension is similar to~\cite{petrovich21actor}. The motion decoder $\mathcal{D}$ relies on the architecture of the transformer decoder with cross attention mechanism, which takes the $L$ number of zero motion tokes as queries, a latent $z \in \mathbb{R}^{n\times d}$ as memory, and finally, generates a human motion sequence $\hat{x}^{1:L}$ with $L$ frames.

According to~\cite{petrovich22temos}, both the latent space $\mathcal{Z}$ and variable durations help the model to produce more diverse motions. To further enhance the latent representation, we leverage a long skip-connection structure for the transformer-based encoder $\mathcal{E}$ and decoder $\mathcal{D}$. We also explore the effectiveness of the latent's dimensions on motion sequences representation in \cref{tab:mr:ablation}. Hence, our VAE models present a stronger motion reconstruction ability and richer diversity ($cf.$ 
\cref{tab:ablation:diffusion} and \cref{tab:ablation:uncon}). We provide more details about the architecture and the training in the supplementary.

\subsection{Motion Latent Diffusion Model}
\label{sec:method:diffusion}

Diffusion probabilistic models~\cite{sohl2015deep} can gradually anneal the noise from a gaussian distribution to a data distribution $p(x)$ by learning the noise prediction from a $T$-length Markov noising process, giving $\{\boldsymbol{x_t}\}^{T}_{t=1}$. It leads to a significant influence in many research domains, such as the most famous image synthesis models~\cite{dhariwal2021diffusion,ho2020denoising,saharia2022image, stable_diffusion}, the density estimation model~\cite{kingma2021variational} and the motion generation models~\cite{mdm2022human,zhang2022motiondiffuse}.
For motion generation, these works train the diffusion models with a transformer-based denoiser $\epsilon_\theta\left(x_t, t\right)$, which anneal the random noise to motion sequence $\{\boldsymbol{\hat{x}_t^{1:N}}\}^{T}_{t=1}$ iteratively.

However, diffusion on raw motion sequences is inefficient and requires exhausting computational resources. Besides, raw motion data from the markless or marker-based motion capture system usually remain high-frequency outliers, which might have a side effect on the diffusion model to learn the actual data distribution. To reduce the computational requirements of the diffusion models on raw motion sequences and improve the synthesized quality, we perform the diffusion process on a representative and low-dimensional motion latent space.

Here, we introduce our denoiser $\epsilon_\theta$. Different from the previous UNet-based architecture~\cite{ronneberger2015u} on the 2D image latent $z_{I}$, we build a transformer-based denoising model with long skip connections~\cite{bao2022all} on the motion latent $z \in \mathbb{R}^{ n\times d}$, which is more suitable for sequential data, like human motion sequences.
The diffusion on latent space is modeled as a Markov nosing process using:
\begin{equation}
q\left(z_t \mid z_{t-1}\right)=\mathcal{N}\left(\sqrt{\alpha_t} z_{t-1},\left(1-\alpha_t\right) I\right).
\end{equation}
where the constant $\alpha_t \in(0,1)$ is a hyper-parameters for sampling.
%
We then use $\{z_t\}^T_{t=0}$ to denote the noising sequence, and $z_{t-1}=\epsilon_\theta\left(z_t, t\right)$ for the $t$-step denoising.
We further focus on the unconditional generation with the simple objective~\cite{ho2020denoising}:
\begin{equation}
    \label[equation]{eq:loss:uncon}
    L_{\text{MLD}}:=\mathbb{E}_{\epsilon, t}\left[\left\|\epsilon-\epsilon_\theta\left(z_t, t\right)\right\|_2^2\right],
\end{equation}
where $\epsilon \sim \mathcal{N}(0,1)$, $z_0 = \mathcal{E}(x^{1:L})$. 
During the training of $\epsilon_\theta$, the encoder $\mathcal{E}$ is frozen to compress motion into $z_0$.
The samples of the diffusion forward process are from the latent distribution $p(z_0)$. 
During the diffusion reverse stage,
$\epsilon_\theta$ first predict $\hat{z}_0$ with $T$ iterative denoising steps, then $\mathcal{D}$ decodes $\hat{z}_0$ to motion results with one forward.

\begin{figure}[t]
    \centering
    \includegraphics[width=0.98\linewidth]{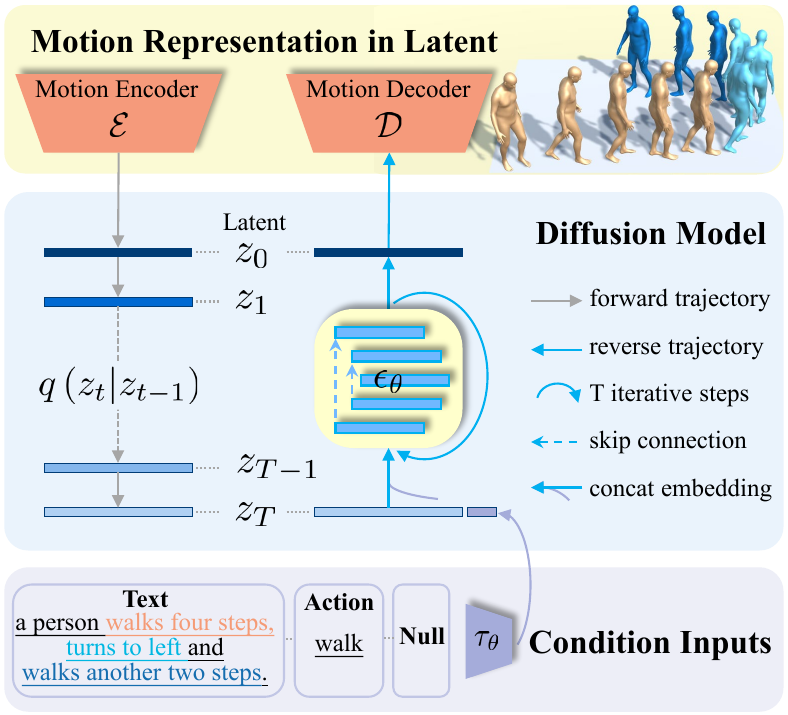}
    \caption{Method overview: MLD consists of a VAE model $\mathcal{V}$ (\cref{sec:method:vae}) and a latent diffusion model $\epsilon_\theta$ (\cref{sec:method:diffusion}) conditioned on text or action embedding $\tau_\theta$ (\cref{sec:method:condition}). We propose two-stage training: first learn $\mathcal{V}$ for \textit{Motion Representations in Latents} and then learn a conditioned denoiser $\epsilon_\theta$ from the diffusion process $q\left(z_t \mid z_{t-1}\right)$. During inference, In practice, the latent diffusion models predict the latent $\hat{z}_0$ from \textit{condition inputs} and then $\mathcal{D}$ decode it to motions efficiently.}
    \label{fig:pipleine}
\end{figure}

\subsection{Conditional Motion Latent Diffusion Model}
\label{sec:method:condition}
Like many other diffusion models~\cite{mdm2022human, stable_diffusion, imagen_saharia2022photorealistic}, our MLD model is also capable of conditional motion generation $\mathcal{G}(c)$ by applying the conditional distribution of $p(z|c)$, such as text~\cite{petrovich22temos,Guo_2022_CVPR_t2m} and action~\cite{petrovich21actor,guo2020action2motion}.
$\mathcal{G}(c)$ is implemented with conditional denoiser $\epsilon_\theta(z_t,t,c)$, which can share a common motion VAE model.
Therefore, for different conditions, only the learning of $\epsilon_\theta(z_t,t,c)$ is necessary.
To address various $c$, the domain encoder $\tau_\theta(c) \in \mathbb{R}^{m \times d}$ for condition embedding benefits the denoiser $\epsilon_\theta(z_t,t,\tau_\theta(c))$.

\begin{figure*}[t]
	\centering
	\includegraphics[width=0.9\linewidth]{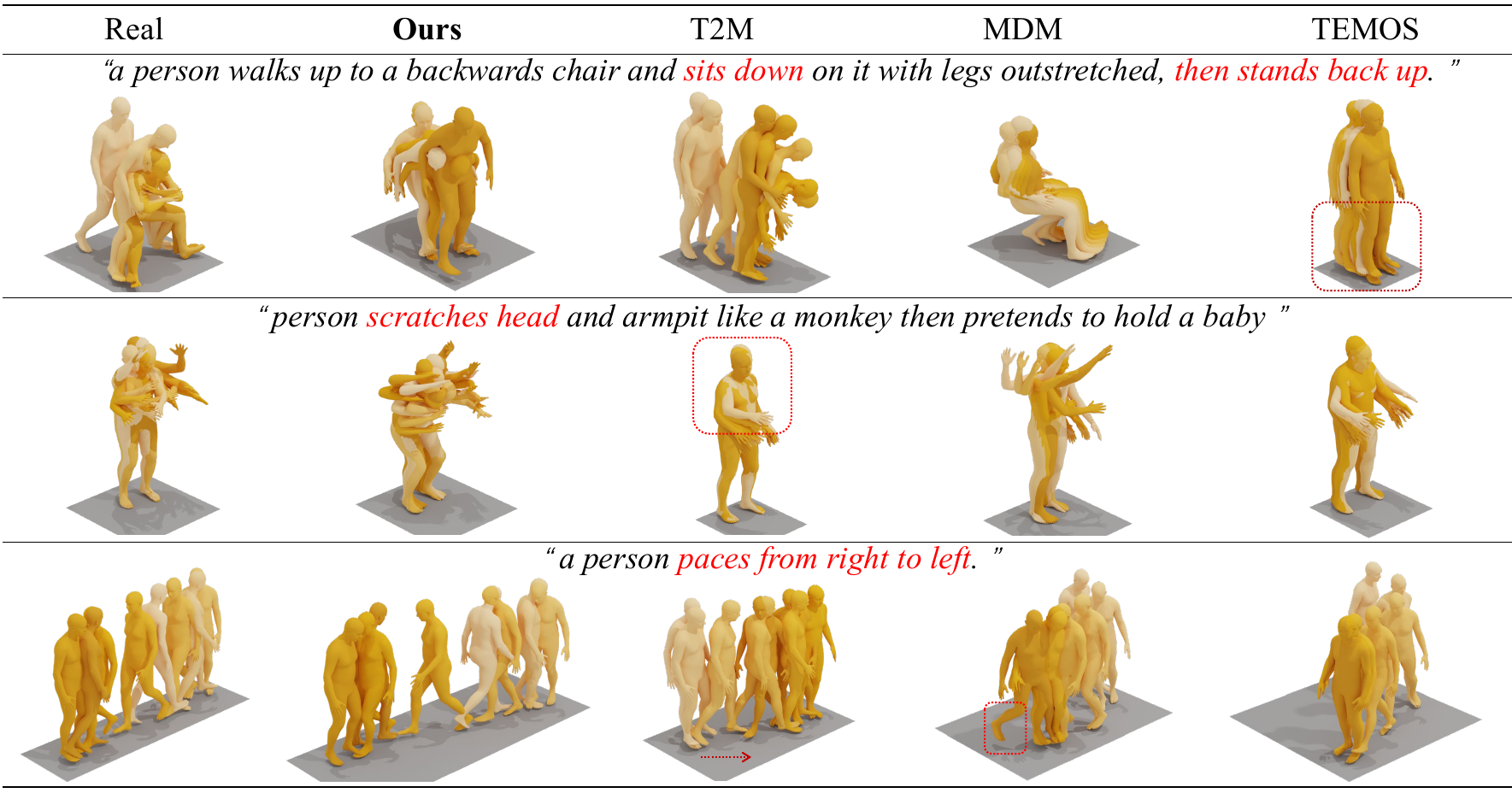}
	\caption{Qualitative comparison of the state-of-the-art methods. We provide the visualized motion results and real references from three text prompts. Under the same training and inference setting on HumanML3D~\cite{Guo_2022_CVPR_humanml3d}, we find that our generations better match the descriptions, but others have downgraded motions or improper semantics.}
	\label{fig:tm:comp}
\end{figure*}

Here we introduce two specific generation tasks, text-to-motion $\mathcal{G}_w: w^{1:N} \mapsto x^{1:L}$ and action-to-motion $\mathcal{G}_a: a \mapsto x^{1:L}$.
Through investigation, CLIP~\cite{radford2021learning} text encoder $\tau_\theta^w(w^{1:N}) \in \mathbb{R}^{1 \times d}$ is employed to map text prompt.
On the other side, we build the learnable embedding for each action category, giving $\tau_\theta^a(a) \in \mathbb{R}^{1 \times d}$.
Injecting these embedded conditions into a transformer-based $\epsilon_\theta$, two effective ways are concatenation and cross-attention, and we figure out the former one seems to be more effective ($cf.$ \cref{sec:ablation:diffusion} and \cite{mdm2022human}) for motion diffusion models.
Thus the conditional objective follows:
\begin{equation}
    \label[equation]{eq:loss:con}
    L_{\text{MLD}}:=\mathbb{E}_{\epsilon, t, c}\left[\left\|\epsilon-\epsilon_\theta\left(z_t, t, \tau_\theta(c)\right)\right\|_2^2\right].
\end{equation}
We freeze $\tau_\theta^w$ as suggested by ~\cite{petrovich22temos} and joint optimize the $\tau_\theta^a$ and $\epsilon_\theta$ via this objective.
In addition, our denoiser $\epsilon_\theta$ is learned with classifier-free diffusion guidance~\cite{ho2022classifier}, which is a trade-off to boost sample quality by reducing diversity in conditional diffusion models.
Specifically, it learns both the conditioned and the unconditioned distribution with 10\% dropout~\cite{imagen_saharia2022photorealistic} of the samples, and we perform a linear combination to in as followed:
\begin{equation}
    \label[equation]{eq:guidance}
    \epsilon^s_\theta(z_t,t,c) = s\epsilon_\theta(z_t,t,c)+ (1-s)\epsilon_\theta(z_t,t,\varnothing)
\end{equation}
Here, $s$ is the guidance scale and $s>1$ can strengthen the effect of guidance.
After the interactive reverse process of the conditional denoising, $\mathcal{D}$ reconstructs the motion from the predicted $\hat{z}_0$ efficiently.
%

\section{Experiments}
\label{experiments}
We provide extensive comparisons to evaluate our models on both quality and efficiency in the following. Firstly, we introduce the datasets settings, evaluation metrics and implementation details (\cref{sec:comp:detail}). 
Importantly, we show the comparisons on multiple datasets for different motion generation tasks respectively, including text-to-motion (\cref{sec:comp:text}), action-to-motion (\cref{sec:comp:action}) and unconditional generation (\cref{sec:comp:uncon}).
More qualitative results, user studies, and details are provided in supplements.

\subsection{Datasets and Evaluation Metrics}
\label{sec:comp:metric}
Conditional motion synthesis can support rich inputs of multi-modal data, and thus multiple datasets are utilized to evaluate MLD.
We briefly introduce these datasets.
%
First is two text-to-motion datasets, HumanML3D and \textbf{KIT}~\cite{Plappert2016kit}, and the latter provides 6,353 textual descriptions for 3,911 motions.
\textbf{HumanML3D}~\cite{Guo_2022_CVPR_humanml3d}, a recent dataset, collects 14,616 motion sequences from \textbf{AMASS}~\cite{AMASS_ICCV2019} and annotates 44,970 sequence-level textual description.
We use its motions, part of the AMASS, to evaluate unconditional task.
%
As suggested by \cite{Guo_2022_CVPR_humanml3d}, we use the redundant motion representation in a combination of joint velocities, positions and rotations which is also used in \cite{mdm2022human,zhang2022motiondiffuse}.
Lastly, action-to-motion task requires action-conditioned motions similar to action recognition datasets.
Thanks to \cite{petrovich21actor}, after the processing,
\textbf{HumanAct12}~\cite{guo2020action2motion} provides 1,191 raw motion sequences and 12 action categories, and \textbf{UESTC}~\cite{ji2018large} provides 24K sequences and 40 action categories.
We rely on these two datasets for action-to-motion evaluation.

\textbf{Evaluation Metrics} summarize in four parts.
%
(a) Motion quality: Frechet Inception Distance (FID) is our principal metric to evaluate the feature distributions between the generated and real motions by feature extractor~\cite{Guo_2022_CVPR_t2m}. 
%
To evaluate reconstruction error of VAEs, we use popular metrics in motion capture~\cite{VIBE_CVPR2020, chen2021sportscap, vonMarcard2018}, 
MPJPE and PAMPJPE~\cite{gower1975generalized} for global/local errors in millimeters, 
Acceleration Error (ACCL) for temporal quality.
(2) Generation diversity: Diversity (DIV) calculates variance through features~\cite{Guo_2022_CVPR_t2m}, while MultiModality (MM) measures the generation diversity within the same text or action input.
(3) Condition matching:
%
Under feature~\cite{Guo_2022_CVPR_t2m} space, 
motion-retrieval precision (R Precision) calculates the text and motion Top 1/2/3 matching accuracy, and Multi-modal Distance (MM Dist) calculates the distance between motions and texts.
For action-to-motion, we use the corresponding action recognition model~\cite{guo2020action2motion}\cite{petrovich21actor} to calculate Accuracy (ACC) for action categories.
%
(4) Time costs: we propose Average Inference Time per Sentence (AITS) measured in seconds to evaluate inference efficiency of diffusion models. ($cf.$ supplements)
%
%




\subsection{Implementation Details}
\label{sec:comp:detail}
For the comparisons, motion transformer encoders $\mathcal{E}$ and decoders $\mathcal{D}$ of our VAE model $\mathcal{V}$ all consist of 9 layers and 4 heads with skip connection by default, as well as the transformer-based denoiser $\epsilon_\theta$ in \cref{sec:method:diffusion}.
The condition embedding $\tau_\theta(c)\in\mathbb{R}^{1,256}$ 
and the latent $z\in\mathbb{R}^{1,256}$ are concatenated for diffusion learning and inference.
We employ a frozen \textit{CLIP-ViT-L-14} model as the text encoder $\tau_\theta^{w}$ for text condition, and a learnable embedding for action condition.
%
We leave the ablation on the components in \cref{sec:ablation}, like the shape of latent, the number of layers, injection of $z_t$ through the cross-attention, and others.
%
%
All our models are trained with the AdamW optimizer using a fixed learning rate of $10^{-4}$. Our mini-batch size is set to 128 during the VAE training stage and 64 during the diffusion training stage separately. Each model was trained for 6K epochs during VAE stage and 3K epochs during diffusion stage.
The number of diffusion steps is 1K during training while 50 during interfering, and the variances $\beta_t$ are scaled linearly from $8.5\times10^{-4}$ to 0.012.
For runtime, training tasks 8 hours for VAEs $\mathcal{V}$ and 4 hours for denoiser $\epsilon_\theta$ on 8 Tesla V100 GPUs, and we test MLD with a single V100 in \cref{sec:runtime}, but it also can run inference on a general computer graphics card, such as RTX 2080/3060.




%
%
%
\begin{table}[t]
\vspace{2pt}
\resizebox{\columnwidth}{!}{%
\begin{tabular}{@{}lccccccc@{}}
\toprule
\multirow{2}{*}{Methods} & \multicolumn{3}{c}{R Precision $\uparrow$}                                                                                                                & \multicolumn{1}{c}{\multirow{2}{*}{FID$\downarrow$}} & \multirow{2}{*}{MM Dist$\downarrow$}              & \multirow{2}{*}{Diversity$\rightarrow$}           & \multirow{2}{*}{MModality$\uparrow$}              \\ \cmidrule(lr){2-4}
              & \multicolumn{1}{c}{Top 1} & \multicolumn{1}{c}{Top 2} & \multicolumn{1}{c}{Top 3} & \multicolumn{1}{c}{}                     &                          &                            &                            \\ \midrule
Real &
  $0.511^{\pm.003}$ &
  $0.703^{\pm.003}$ &
  $0.797^{\pm.002}$ &
  $0.002^{\pm.000}$ &
  $2.974^{\pm.008}$ &
  $9.503^{\pm.065}$ &
  \multicolumn{1}{c}{-}
  \\ \midrule
Seq2Seq \cite{plappert2018learning} &
  $0.180^{\pm.002}$ &
  $0.300^{\pm.002}$ &
  $0.396^{\pm.002}$ &
  $11.75^{\pm.035}$ &
  $5.529^{\pm.007}$ &
  $6.223^{\pm.061}$ &
  \multicolumn{1}{c}{-} \\
LJ2P \cite{ahuja2019language2pose}&
  $0.246^{\pm.001}$ &
  $0.387^{\pm.002}$ &
  $0.486^{\pm.002}$ &
  $11.02^{\pm.046}$ &
  $5.296^{\pm.008}$ &
  $7.676^{\pm.058}$ &
  \multicolumn{1}{c}{-} \\
T2G\cite{bhattacharya2021text2gestures} &
  $0.165^{\pm.001}$ &
  $0.267^{\pm.002}$ &
  $0.345^{\pm.002}$ &
  $7.664^{\pm.030}$ &
  $6.030^{\pm.008}$ &
  $6.409^{\pm.071}$ &
  \multicolumn{1}{c}{-} \\
Hier \cite{ghosh2021synthesis}&
    $0.301^{\pm.002}$ &
    $0.425^{\pm.002}$ &
    $0.552^{\pm.004}$ &
    $6.532^{\pm.024}$ &
    $5.012^{\pm.018}$ &
    $8.332^{\pm.042}$ &
    \multicolumn{1}{c}{-} \\
TEMOS \cite{petrovich22temos}&
  $0.424^{\pm.002}$ &
  $0.612^{\pm.002}$ &
  $0.722^{\pm.002}$ &
  $3.734^{\pm.028}$ &
  $3.703^{\pm.008}$ &
  $8.973^{\pm.071}$ &
  $0.368^{\pm.018}$ \\
T2M \cite{Guo_2022_CVPR_t2m}&
  $0.457^{\pm.002}$ &
  $0.639^{\pm.003}$ &
  $0.740^{\pm.003}$ &
  $1.067^{\pm.002}$ &
  $3.340^{\pm.008}$ &
  $9.188^{\pm.002}$ &
  $2.090^{\pm.083}$ \\
MDM \cite{mdm2022human}&
  $0.320^{\pm.005}$ &
  $0.498^{\pm.004}$ &
  $0.611^{\pm.007}$ &
  $\underline{0.544}^{\pm.044}$ &
  $5.566^{\pm.027}$ &
  $\boldsymbol{9.559}^{\pm.086}$ &
  $\boldsymbol{2.799}^{\pm.072}$ \\
 MotionDiffuse \cite{zhang2022motiondiffuse} &
  $\boldsymbol{0.491}^{\pm.001}$ &
  $\boldsymbol{0.681}^{\pm.001}$ &
  $\boldsymbol{0.782}^{\pm.001}$ &
  $0.630^{\pm.001}$ &
  $\boldsymbol{3.113}^{\pm.001}$ &
  $\underline{9.410}^{\pm.049}$ &
  $1.553^{\pm.042}$ \\
 \midrule
MLD (Ours) &
  $\underline{0.481}^{\pm.003}$ &
  $\underline{0.673}^{\pm.003}$ &
  $\underline{0.772}^{\pm.002}$ &
  $\boldsymbol{0.473}^{\pm.013}$ &
  $\underline{3.196}^{\pm.010}$ &
  $9.724^{\pm.082}$ &
  $\underline{2.413}^{\pm.079}$ 
   \\ \bottomrule
\end{tabular}%
}
\vspace{-6pt}
\caption{Comparison of text-conditional motion synthesis on HumanML3D~\cite{Guo_2022_CVPR_humanml3d} dataset. These metrics are evaluated by the motion encoder from \cite{Guo_2022_CVPR_t2m}. Empty MModality indicates the non-diverse generation methods. We employ real motion as a reference and sort all methods by descending FIDs. The right arrow $\rightarrow$ means the closer to real motion the better. \textbf{Bold} and \underline{underline} indicate the best and the second best result.}
\label{tab:tm:comp:humanml3d}
\end{table}

\begin{table}[t]
\resizebox{\columnwidth}{!}{%
\begin{tabular}{@{}lcccccccc@{}}
\toprule
\multirow{2}{*}{Methods} & \multicolumn{3}{c}{R Precision $\uparrow$}                                                                                                                & \multicolumn{1}{c}{\multirow{2}{*}{FID$\downarrow$}} & \multirow{2}{*}{MM Dist$\downarrow$}              & \multirow{2}{*}{Diversity$\rightarrow$}           & \multirow{2}{*}{MModality$\uparrow$}              \\ \cmidrule(lr){2-4}
              & \multicolumn{1}{c}{Top 1} & \multicolumn{1}{c}{Top 2} & \multicolumn{1}{c}{Top 3} & \multicolumn{1}{c}{}                     &                          &                            &                            \\ \midrule
Real &
  $0.424^{\pm.005}$ &
  $0.649^{\pm.006}$ &
  $0.779^{\pm.006}$ &
  $0.031^{\pm.004}$ &
  $2.788^{\pm.012}$ &
  $11.08^{\pm.097}$ &
  \multicolumn{1}{c}{-}
  \\ \midrule
Seq2Seq\cite{plappert2018learning} &
  $0.103^{\pm.003}$ &
  $0.178^{\pm.005}$ &
  $0.241^{\pm.006}$ &
  $24.86^{\pm.348}$ &
  $7.960^{\pm.031}$ &
  $6.744^{\pm.106}$ &
  \multicolumn{1}{c}{-} \\
T2G\cite{bhattacharya2021text2gestures} &
  $0.156^{\pm.004}$ &
  $0.255^{\pm.004}$ &
  $0.338^{\pm.005}$ &
  $12.12^{\pm.183}$ &
  $6.964^{\pm.029}$ &
  $9.334^{\pm.079}$ &
  \multicolumn{1}{c}{-} \\
LJ2P \cite{ahuja2019language2pose}&
  $0.221^{\pm.005}$ &
  $0.373^{\pm.004}$ &
  $0.483^{\pm.005}$ &
  $6.545^{\pm.072}$ &
  $5.147^{\pm.030}$ &
  $9.073^{\pm.100}$ &
  \multicolumn{1}{c}{-} \\
Hier \cite{ghosh2021synthesis}&
  $0.255^{\pm.006}$ &
  $0.432^{\pm.007}$ &
  $0.531^{\pm.007}$ &
  $5.203^{\pm.107}$ &
  $4.986^{\pm.027}$ &
  $9.563^{\pm.072}$ &
  $2.090^{\pm.083}$ \\
TEMOS \cite{petrovich22temos}&
    $0.353^{\pm.006}$ & 
    $0.561^{\pm.007}$ & 
    $0.687^{\pm.005}$ & 
    $3.717^{\pm.051}$ & 
    $3.417^{\pm.019}$ & 
    $10.84^{\pm.100}$ & 
    $0.532^{\pm.034}$ \\
T2M \cite{Guo_2022_CVPR_t2m}&
  $0.370^{\pm.005}$ &
  $0.569^{\pm.007}$ &
  $0.693^{\pm.007}$ &
  $2.770^{\pm.109}$ &
  $3.401^{\pm.008}$ &
  $\underline{10.91}^{\pm.119}$ &
  $1.482^{\pm.065}$ \\
MDM \cite{mdm2022human}&
  $0.164^{\pm.004}$ &
  $0.291^{\pm.004}$ &
  $0.396^{\pm.004}$ &
  $\underline{0.497}^{\pm.021}$ &
  $9.191^{\pm.022}$ &
  $10.85^{\pm.109}$ &
  $\underline{1.907}^{\pm.214}$ \\
MotionDiffuse \cite{zhang2022motiondiffuse} &
$\boldsymbol{0.417}^{\pm.004}$ &
$\boldsymbol{0.621}^{\pm.004}$ &
$\boldsymbol{0.739}^{\pm.004}$ &
$1.954^{\pm.062}$ &
$\boldsymbol{2.958}^{\pm.005}$ &
$\boldsymbol{11.10}^{\pm.143}$ &
$0.730^{\pm.013}$ \\
\midrule
MLD (Ours) &
$\underline{0.390}^{\pm.008}$ & 
$\underline{0.609}^{\pm.008}$ & 
$\underline{0.734}^{\pm.007}$ & 
$\boldsymbol{0.404}^{\pm.027}$ & 
$\underline{3.204}^{\pm.027}$ & 
$10.80^{\pm.117}$ & 
$\boldsymbol{2.192}^{\pm.071}$
   \\ \bottomrule
\end{tabular}%
}
\vspace{-8pt}
\caption{We involve KIT~\cite{Plappert2016kit} dataset and evaluate the SOTA methods on the text-to-motion task. ($cf.$ \cref{tab:tm:comp:humanml3d} for metrics details)}
\label{tab:tm:comp:kit}
\end{table}

\begin{figure*}[t] 
	\centering 
	\includegraphics[width=0.8\linewidth]{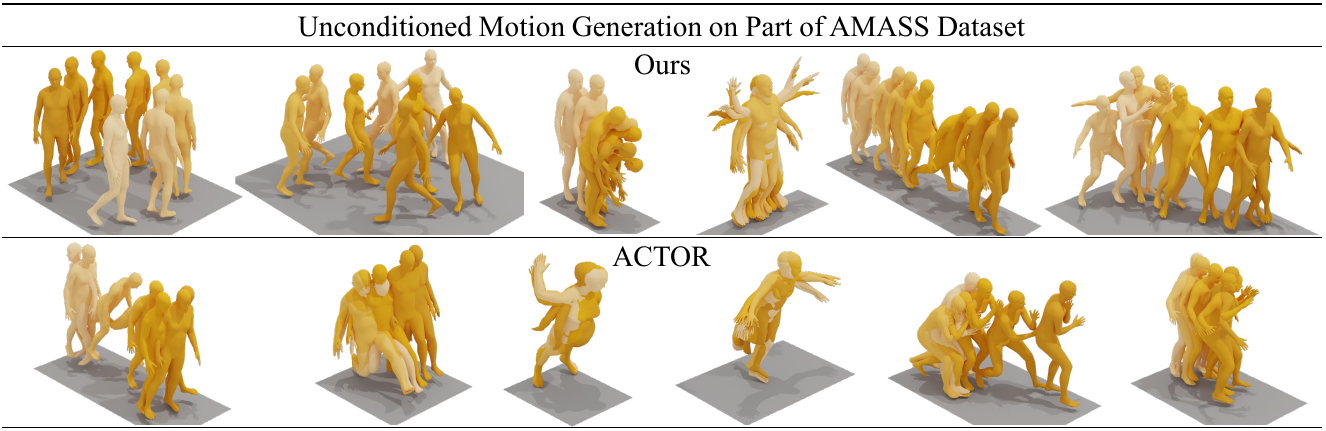} 
        \vspace{-8pt} 
	\caption{Qualitative comparison of unconditioned motion generation. Samples for our MLD and ACTOR~\cite{petrovich21actor} trained on a split of AMASS~\cite{AMASS_ICCV2019} dataset, the motion part of \cite{Guo_2022_CVPR_humanml3d}. More samples in the supplements.} 
	\label{fig:uncon} 
	\vspace{-8pt} 
\end{figure*}

\subsection{Comparisons on Text-to-motion}
\label{sec:comp:text}
By introducing motion latent diffusion models based on text input $w^{1:N}$, we open up the exploration of conditional motion generation.
We train a 25M parameter \textit{MLD-1} conditioned on the language prompt and employ the frozen CLIP~\cite{radford2021learning} model as $\tau_\theta^{w}$ to encode the text to projected pooled output, giving $w_{clip}^{1}\in \mathbb{R}^{1,256}$.
We evaluate state-of-the-art methods on HumanML3D and KIT with suggested metrics~\cite{Guo_2022_CVPR_humanml3d} under the $95\%$ confidence interval from 20 times running.
Most results are borrowed from their own paper or the benchmark in \cite{chuan2022tm2t}, except TEMOS~\cite{petrovich22temos}.
We train it with the proposed default model setting on two datasets to uniform the evaluation metrics.
Besides, the deterministic methods~\cite{plappert2018learning, bhattacharya2021text2gestures, ahuja2019language2pose} can not generate diverse results from one input and thus we leave their MModality metrics empty.
\cref{tab:tm:comp:humanml3d} and \cref{tab:tm:comp:kit} summarize the comparisons results.
We achieve the best FID, R Precision and MM Dist on HumanML3D and KIT, outperforming previous cross-modal models as well as motion diffusion models.
It indicates high-quality motion and high text prompt matching, as also shown in \cref{fig:tm:comp}.
Our generated results correctly match the text prompt while maintaining a rich diversity of generated motions.
%
%
%


\begin{table}[h]
\resizebox{\columnwidth}{!}{%
\begin{tabular}{@{}lccccccccc@{}}
\toprule
\multirow{2}{*}{Methods} & \multicolumn{5}{c}{UESTC}                                               & \multicolumn{4}{c}{HumanAct12}                          \\ \cmidrule(lr){2-6} \cmidrule(l){7-10} 
                         & $\text{FID}_{\text{train}}\downarrow $            & $\text{FID}_{\text{test}}\downarrow$             & ACC$\uparrow$ & DIV$\rightarrow$ & MM$\rightarrow$ &   \multicolumn{1}{c}{$\text{FID}_{\text{train}}\downarrow$} & ACC $\uparrow$& DIV$\rightarrow$ & MM$\rightarrow$ \\ 
\toprule
                         
Real &
  $2.92^{\pm.26}$ &
  $2.79^{\pm.29}$ &
  $0.988^{\pm.001}$ &
  $33.34^{\pm.320}$ &
  $14.16^{\pm.06}$ &
  $0.020^{\pm.010}$ &
  $0.997^{\pm.001}$ &
  $6.850^{\pm.050}$ &
  $2.450^{\pm.040}$ \\
  \midrule
ACTOR \cite{petrovich21actor}&
  $20.5^{\pm2.3}$ &
  $23.43^{\pm2.20}$ &
  $0.911^{\pm.003}$ &
  $31.96^{\pm.33}$ &
  $14.52^{\pm.09}$ &
  $0.120^{\pm.000}$ &
  $0.955^{\pm.008}$ &
  $6.840^{\pm.030}$ &
  $2.530^{\pm.020}$ \\
INR \cite{cervantes2022implicit}&
  $\boldsymbol{9.55}^{\pm.06}$ &
  $15.00^{\pm.09}$ &
  $0.941^{\pm.001}$ &
  $31.59^{\pm.19}$ &
  $14.68^{\pm.07}$ &
  $0.088^{\pm.004}$ &
  $0.973^{\pm.001}$ &
  $6.881^{\pm.048}$ &
  $2.569^{\pm.040}$ \\
MDM \cite{mdm2022human}&
  $9.98^{\pm1.33}$ &
  $\boldsymbol{12.81}^{\pm1.46}$ &
  $0.950^{\pm.000}$ &
  $33.02^{\pm.28}$ &
  $\boldsymbol{14.26}^{\pm.12}$ &
  $0.100^{\pm.000}$ &
  $\boldsymbol{0.990}^{\pm.000}$ &
  $6.680^{\pm.050}$ &
  $\boldsymbol{2.520}^{\pm.010}$\\
  \midrule
MLD (Ours) &

$12.89^{\pm.109}$ &
$15.79^{\pm.079}$ &
$\boldsymbol{0.954}^{\pm.001}$ &
$\boldsymbol{33.52}^{\pm.14}$ &
$13.57^{\pm.06}$ &
$\boldsymbol{0.077}^{\pm.004}$&
$0.964^{\pm.002}$&
$\boldsymbol{6.831}^{\pm.050}$&
$2.824^{\pm.038}$
    
\\ \bottomrule
\end{tabular}%
}
\vspace{-8pt}
\caption{Comparison of action-conditional motion synthesis on UESTC~\cite{ji2018large} and HumanAct12~\cite{guo2020action2motion} dataset: $\text{FID}_\text{train}$, $\text{FID}_\text{train}$ indicate the evaluated splits. Accuracy (ACC) for action recognition. Diversity (DIV), MModality (MM) for generated motion diversity within each action label.}
\label{tab:comp:action}
\end{table}

\subsection{Comparisons on Action-to-motion}
\label{sec:comp:action}
The action-conditioned task is given an input action label to generate relevant motion sequences.
We compare with ACTOR~\cite{petrovich21actor}, INR~\cite{cervantes2022implicit} and MDM~\cite{mdm2022human}.
ACTOR and INR are transformer-based VAE models and focus on the action-conditioned task, and MDM is a diffusion model using the same learnable action embedding module as ours.
We still provide 20 evaluations as introduced and report FID scores on the training set and test set like ~\cite{petrovich21actor} for comparison.
\cref{tab:comp:action} shows the comparison on two datasets, UESTC~\cite{ji2018large} and HumanAct12~\cite{guo2020action2motion}.
MLD achieves state-of-the-art accuracy and diversity on UESTC and competitive results on HumanAct12, indicating that diffusion models in motion latent can also benefit action-conditioned generation task.

\begin{figure}[t]
    \centering
    \includegraphics[width=0.80\linewidth]{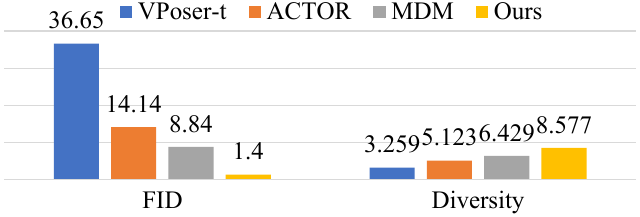}
    \caption{Comparison of unconditional motion generation on part of AMASS~\cite{AMASS_ICCV2019} dataset with the state-of-the-art methods. We provide both FID and Diversity to evaluate generated motions. }
    \label{fig:comp:uncondition}
\end{figure}

\subsection{Comparisons on Unconditional Generation}
\label{sec:comp:uncon}
We then evaluate the generation effect of MLD by introducing unconditional task on motions of HumanML3D~\cite{Guo_2022_CVPR_humanml3d}, actually part of AMASS~\cite{AMASS_ICCV2019}.
MLD supports two manners for unconditional generation, latent sampling ($cf.$ \cref{sec:exp:vae}) and diffusion sampling.
Here we focus on the evaluation of the latter and employ FID and
Diversity for motion quality and diversity.
%
%
With the same process on training and evaluations on the part of AMASS~\cite{AMASS_ICCV2019} data, we provide real motion, ACTOR~\cite{petrovich21actor}, and VPoser-t~\cite{vposer_SMPL-X:2019} and MDM~\cite{mdm2022human} as our comparison baselines,
We employ the transformer VAE from ACTOR, then follow TEMOS~\cite{petrovich22temos} to make it class-agnostic and set 6 heads/layers for transformers, $10^{-4}$ as learning rate.
To perform the temporal-based task, the input of VPoser-t is modified as a motion of fixed length.
MDM also supports this task, thus we fine-tune and evaluate their provided model.
\cref{fig:comp:uncondition} reports that MLD has the best motion generation quality and diversity.

\section{Ablation Studies}
\label{sec:ablation}
MLD comprises a motion VAE model $\mathcal{V}$ and latent diffusion models $\epsilon_\theta$, and both influence its effect.
We first focus on $\mathcal{V}$ to evaluate its components with generation and reconstruction metrics.
Based on these $\mathcal{V}$, we evaluate MLDs in diffusion learning aiming at text-to-motion and unconditional synthesis,
and then report time costs on inference.

\begin{table}[t]
\resizebox{\columnwidth}{!}{%
\begin{tabular}{@{}lccclccl@{}}
\toprule
\multirow{2}{*}{Method} & \multicolumn{3}{c}{Reconstruction}                            &  & \multicolumn{2}{c}{Generation}        \\ \cmidrule(lr){2-4} \cmidrule(l){6-7} 
                        & MPJPE $\downarrow$ & PAMPJPE$\downarrow$ & ACCL$\downarrow$ &  & FID$\downarrow$& DIV$\rightarrow$\\ \midrule
Real  &  \multicolumn{1}{c}{-} & \multicolumn{1}{c}{-}& \multicolumn{1}{c}{-} & &  $0.002\hidden{^{\pm.000}}$&	$9.503\hidden{^{\pm.065}}$     \\ \midrule
VPoser-t~\cite{vposer_SMPL-X:2019}  & $75.6$ & $48.6$& $9.3$ &     & $1.430\hidden{^{\pm.002}}$ & $8.336\hidden{^{\pm.099}}$   \\
ACTOR~\cite{petrovich21actor}     & $65.3$ & $41.0$& $7.0$ &     & $0.341\hidden{^{\pm.002}}$ & $9.569\hidden{^{\pm.099}}$ \\
 \midrule
\textbf{Ours-7} ($\mathcal{V}$,skip,9 layers) & $\boldsymbol{14.7}$ & $ \boldsymbol{8.9}$ & $\boldsymbol{5.1}$ &  & $\boldsymbol{0.017}\hidden{^{\pm.000}}$ &  $\boldsymbol{9.554}\hidden{^{\pm.062}}$  \\
\toprule
\toprule
Ours-1 ($z$,$\mathbb{R}^{1\times256}$) & $54.4$ & $41.6$ & $8.3$ &  & $0.247\hidden{^{\pm.001}}$ & $9.630\hidden{^{\pm.081}}$ \\
Ours-2 ($z$,$\mathbb{R}^{2\times256}$) & $51.8$ & $37.8$ & $8.3$ &  & $0.166\hidden{^{\pm.001}}$ &  $9.626\hidden{^{\pm.087}}$ \\
Ours-5 ($z$,$\mathbb{R}^{5\times256}$) & $24.3$ & $14.7$ & $5.8$ &  & $0.043\hidden{^{\pm.000}}$ & $9.593\hidden{^{\pm.084}}$ \\
Ours-7 ($z$,$\mathbb{R}^{7\times256}$) & $14.7$ & $ 8.9$ & $5.1$ &  & $0.017\hidden{^{\pm.000}}$ &  $9.554\hidden{^{\pm.062}}$\\
Ours-10 ($z$,$\mathbb{R}^{10\times256}$) & $17.3$ & $11.5$ &  $5.8$  && $0.025\hidden{^{\pm.000}}$ & $9.589\hidden{^{\pm.090}}$  \\
\midrule
Ours-7 ($\mathcal{V}$,w/ skip) & $14.7$ & $ 8.9$ & $5.1$ &  & $0.017\hidden{^{\pm.000}}$ & $9.554\hidden{^{\pm.062}}$   \\
Ours-7 ($\mathcal{V}$, w/o skip) &$18.5$ & $10.4$ & $5.6$ && $0.027\hidden{^{\pm.000}}$ & $9.528\hidden{^{\pm.067}}$ &  \\
\midrule
Ours-7 ($\mathcal{V}$, 7 layers) & $16.0$ & $10.2$ & $5.3$ && $0.022\hidden{^{\pm.000}}$ &  $9.593\hidden{^{\pm.075}}$ \\
Ours-7 ($\mathcal{V}$, 9 layers) &$14.7$ & $ 8.9$ & $5.1$ &  & $0.017\hidden{^{\pm.000}}$ & $9.554\hidden{^{\pm.062}}$  \\
Ours-7 ($\mathcal{V}$, 11 layers) & $17.2$& $11.2$ & $5.4$ && $0.021\hidden{^{\pm.000}}$ &  $9.533\hidden{^{\pm.092}}$ & \\ \bottomrule
\end{tabular}%
}
\vspace{-2pt}
\caption{Evaluation of our VAE models $\mathcal{V}$ on the motion part of HumanML3D~\cite{Guo_2022_CVPR_humanml3d} dataset: MPJPE and PAMPJPE are measured in millimeter. ACCL indicates acceleration error. We evaluate FID and DIV the same as \cref{tab:tm:comp:humanml3d}. From top to down, we propose real reference, VPoser-t~\cite{vposer_SMPL-X:2019} and ACTOR~\cite{petrovich21actor} as baselines, the evaluation on latent $z\in\mathbb{R}^{i\times256}$, with (w/) or without (w/o) skip connection, $\mathcal{V}$ with different number of transformer layers.}
\label{tab:mr:ablation}
\vspace{-9pt}
\end{table}

\label{sec:exp:vae}
\textbf{Effectiveness of Latents in Motion Sequences Representation.}
We first ablate several components of our VAE models $\mathcal{V}$ in a controlled setup, studying the shape of latent $z$, skip connection, and the number of transformer layers, as shown in \cref{tab:mr:ablation}.
The most important variable of MLD, the latent vector $z$, is a bridge between $\mathcal{V}$ and diffusion models $\tau_{\theta}$.
We lock the pose $x^i$ (one frame of motion) embedding dimensionality to 256, which is the same as \cite{petrovich22temos} 
, and explore $z\in\mathbb{R}^{i\times 256}$, giving \text{MLD}-$i$.
We then evaluate the skip connection and transformer layers on the best MLD-7.
All comparison baselines, including ACTOR~\cite{petrovich21actor} and VPoser-t~\cite{vposer_SMPL-X:2019}, follow the same training and evaluation with our proposed MLD. Since the original VPoser can only handle single frame pose, we modified it to a sequential manner with a fixed length. The results in~\cref{tab:mr:ablation} demonstrate the effectiveness of our proposed VAEs over others in the motion sequences representation.

\begin{table}[t]
\resizebox{\columnwidth}{!}{%
\begin{tabular}{@{}lccccc@{}}
\toprule
\multirow{2}{*}{Models} & \multicolumn{1}{c}{R Precision} & \multicolumn{1}{c}{\multirow{2}{*}{FID$\downarrow$}} & \multirow{2}{*}{MM Dist.$\downarrow$} & \multirow{2}{*}{Diversity$\rightarrow$} & \multirow{2}{*}{MModality$\uparrow$} \\
                           & \multicolumn{1}{c}{Top 3$\uparrow$}       & \multicolumn{1}{c}{}                     &                          &                            &     \\ \toprule
Real &
  $0.797^{\pm.002}$ &
  $0.002^{\pm.000}$ &
  $2.974^{\pm.008}$ &
  $9.503^{\pm.065}$ &
  \multicolumn{1}{c}{-}
  \\ \midrule
MLD-1 ($z$, $\mathbb{R}^{1\times256}$) & 
  $0.772^{\pm.002}$ & $0.473^{\pm.013}$ & $3.196^{\pm.010}$ & $9.724^{\pm.082}$ & $2.413^{\pm.079}$ \\
MLD-2 ($z$, $\mathbb{R}^{2\times256}$) & $0.727^{\pm.003}$&		$0.585^{\pm.015}$ &	$3.448^{\pm.011}$ &	$9.084^{\pm.081}$ &	$\boldsymbol{2.725}^{\pm.093}$   \\
MLD-5 ($z$, $\mathbb{R}^{5\times256}$)& $0.722^{\pm.003}$ & $1.554^{\pm.019}$ & $3.511^{\pm.008}$ & $8.424^{\pm.081}$ & $2.542^{\pm.080}$   \\

MLD-7 ($z$, $\mathbb{R}^{7\times256}$)  & $0.731^{\pm.002}$ & $1.011{^{\pm.019}}$ & $3.415{^{\pm.008}}$ & $8.736{^{\pm.064}}$ & $2.463{^{\pm.089}}$ \\
MLD-10 ($z$, $\mathbb{R}^{10\times256}$) & $0.703^{\pm.003}$ & $1.716^{\pm.027}$ & $3.616^{\pm.012}$ & $8.606^{\pm.067}$ & $2.604^{\pm.087}$
\\ \midrule
MLD-1 ($\epsilon_\theta$, cross-att) &$0.592{^{\pm.004}}$ & $1.922{^{\pm.041}}$ & $4.480{^{\pm.015}}$ & $8.598{^{\pm.088}}$ & $3.768{^{\pm.126}}$\\
MLD-1 ($\epsilon_\theta$, concat) & $0.772{^{\pm.002}}$ & $0.473{^{\pm.013}}$ & $3.196{^{\pm.010}}$ & $9.724{^{\pm.082}}$ & $2.413{^{\pm.079}}$ \\
\midrule
MLD-1 ($\epsilon_\theta$, w/o skip) & $0.749{^{\pm.003}}$ & $0.784{^{\pm.015}}$ & $3.363{^{\pm.010}}$ & $9.568{^{\pm.093}}$ & $2.597{^{\pm.098}}$ \\
MLD-1 ($\epsilon_\theta$, w/ skip) & $0.772{^{\pm.002}}$  & $0.473{^{\pm.013}}$ & $3.196{^{\pm.010}}$ & $9.724{^{\pm.082}}$ & $2.413{^{\pm.079}}$ \\
\midrule
MLD-1 ($\epsilon_\theta$, 5 layers) & $0.760{^{\pm.002}}$ & $\boldsymbol{0.314}{^{\pm.010}}$ & $3.259{^{\pm.009}}$ & $9.706{^{\pm.072}}$ & $2.635{^{\pm.085}}$ \\ 
MLD-1 ($\epsilon_\theta$, 7 layers) & $0.771{^{\pm.003}}$ & $0.349{^{\pm.012}}$ & $3.199{^{\pm.012}}$ & $\boldsymbol{9.624}{^{\pm.062}}$ & $2.504{^{\pm.088}}$  \\
MLD-1 ($\epsilon_\theta$, 9 layers) & $\boldsymbol{0.772}{^{\pm.002}}$  & $0.473{^{\pm.013}}$ & $\boldsymbol{3.196}{^{\pm.010}}$ & $9.724{^{\pm.082}}$ & $2.413{^{\pm.079}}$  \\
MLD-1 ($\epsilon_\theta$, 11 layers) & $0.771{^{\pm.003}}$ & $0.402{^{\pm.011}}$ & $3.203{^{\pm.013}}$ & $9.876{^{\pm.088}}$ & $2.478{^{\pm.076}}$ \\

\bottomrule
\end{tabular}%
}
\vspace{-8pt}
\caption{Evaluation of text-based motion synthesis on HumanML3D~\cite{Guo_2022_CVPR_humanml3d}: we use metrics in \cref{tab:tm:comp:humanml3d} and provides real reference, the evaluation on latent $z$ ($cf.$ $\mathcal{V}$ in \cref{tab:mr:ablation}), cross-attention or concatenation with conditions $\tau_\theta$, with (w/) or without (w/o) skip connection, $\epsilon_\theta$ with different number of transformer layers.}
\vspace{-4pt}
\label{tab:ablation:diffusion}
\end{table}

\begin{table}[t]
\centering
\resizebox{\columnwidth}{!}{%
\begin{tabular}{@{}lcclccc@{}}
\toprule
Methods              & {FID$\downarrow$} 
& {Diversity$\rightarrow$} & Methods  & {FID$\downarrow$} 
& {Diversity$\rightarrow$}                           
\\ \toprule
Real & \multicolumn{1}{c}{-} 
& 9.503 &Real 
& \multicolumn{1}{c}{-} 
& 9.503
\\ \midrule
VPoser-t \cite{vposer_SMPL-X:2019} & $36.65$ 
& $3.259$ &MLD-1 ($z$,$\mathbb{R}^{1\times256}$) & $\boldsymbol{1.055}$ 
&$\boldsymbol{8.577}$ 
\\
ACTOR \cite{petrovich21actor} &$14.14$  
& $5.123$ &MLD-2 ($z$,$\mathbb{R}^{1\times256}$) & $4.408$ 
&$7.420$
\\
MDM \cite{mdm2022human}& 8.848 & 6.429 & MLD-5 ($z$,$\mathbb{R}^{5\times256}$) 
& $7.829$ 
&$6.247$
\\
MLD-1 ($\epsilon_\theta$, w/o skip)  & $2.575$ 
&$7.566$ &MLD-7 ($z$,$\mathbb{R}^{7\times256}$)   & $7.614$ 
&$6.233$
\\
MLD-1 ($\epsilon_\theta$, w/ skip)  & $\boldsymbol{1.055}$ 
&$\boldsymbol{8.577}$   &MLD-10 ($z$,$\mathbb{R}^{10\times256}$)  
& $9.624$ 
&$6.194$
\\ 
 \bottomrule
\end{tabular}%
}
\vspace{-8pt}
\caption{Evaluation of unconditional motion generation. From left to right, we evaluate the denoiser $\epsilon_\theta$ with (w/) or without (w/o) skip connection and  the latent $z$ ($cf. \mathcal{V}$ in  \cref{tab:mr:ablation}) }
\label{tab:ablation:uncon}
\vspace{-10pt}
\end{table}


\label{sec:ablation:diffusion}
\textbf{Effectiveness of Latents in Motion Latent-based Diffusion Models.}
In \cref{tab:ablation:diffusion}, we select the text-to-motion task as our focus and evaluate latent diffusion models $\epsilon_\theta$, using the similar metrics in \cref{tab:tm:comp:humanml3d}.
MLD-$i$ denotes the shape of latent $z\in\mathbb{R}^{i\times256}$.
Importantly, MLD-1, using the smallest latent, wins the best performance in most metrics.
%
%
After that, the evaluation on the components of $\epsilon_\theta$ is provided, \textit{cross-att} and \textit{concate} represent the cross-attention or concatenation for condition embedding $\epsilon_\theta(c)$.
Interestingly, MDM~\cite{mdm2022human} also reports the encoder design by concatenating embedding is better.
We find that skip connection, which is important for images~\cite{bao2022all, stable_diffusion}, also provided significant improvement in motion latent diffusion models, but MLDs using different numbers of layers in $\epsilon_\theta$ achieve similar effects on this dataset.
We then evaluate the generation of MLD by diffusion sampling, different from the generation in $\mathcal{V}$ by latent sampling ($cf.$ \cref{tab:mr:ablation}).
As shown in \cref{tab:ablation:uncon}, the MLD using the smallest latent and skip connection outperforms others.
The evaluation of how different language models influence MLDs and the details of latent/diffusion sampling are provided in supplements.

\begin{figure}[t]
    \centering
    \includegraphics[width=0.95\linewidth]{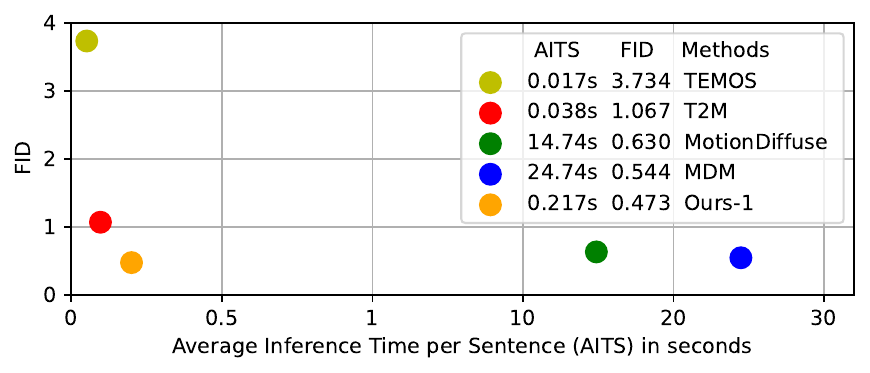}
    \caption{Comparison of the inference time costs on text-to-motion. 
    We calculate AITS on the test set of HumanML3D~\cite{Guo_2022_CVPR_humanml3d} without model or data loading parts. All tests are performed on the same Tesla V100. The closer the model is to the origin the better.}
    \label{fig:comp:time}
\end{figure}

\label{sec:runtime}
\textbf{Inference time.} While diffusion models lead to significant improvements, one notable limitation of motion diffusion models~\cite{mdm2022human, zhang2022motiondiffuse} is the long inference time.
In Sec. C, we adopted Denoising diffusion implicit models (DDIM)~\cite{song2020denoising} to provide a detailed evaluation of the inference time, floating-point operations (FLOPs), and FID.
As shown in \cref{fig:comp:time},
MDM~\cite{mdm2022human} requires 24.74 seconds for average inference and up to a minute for maximum inference on a single V100.
Compared to them, our MLD needs less computational overhead and achieves higher performance with two orders of magnitude faster speed.




\section{Disscusion}
As the  trial to explore conditional motion generation with motion latent diffusion models, the proposed MLD still owns limitations as follows.
First, same as most motion generation methods, our method can generate arbitrary length results but still under the max-length in the dataset.
It's interesting to model a non-stop human motion in temporal consistency.
Besides, MLD focuses on articulated human bodies, while there is also other work on faces~\cite{karras2017audio, cao2018sparse}, hands~\cite{romero2022embodied, li2022nimble, li2021piano} and even animal~\cite{Rueeg:CVPR:2022, Zuffi:CVPR:2018} motion.

We propose a motion latent-based diffusion model to generate plausible human motion sequences conforming to the action classes or natural language descriptions. Compared to the compatible cross-modal latent space-based method, our MLD could produce more diverse and plausible human motion sequences; Compared to the previous diffusion-based methods on raw motion sequences, our MLD needs less computational overhead, with two orders of magnitude faster. Extensive experiments on various human motion generation tasks demonstrate the effectiveness and efficiency of our proposed MLD.



\section{Acknowledgements}

This work is supported by Zhejiang Lab Project (No. 2021KH0AB05) and Shanghai Natural Science Foundation (No. 23ZR1402900).

{\small
    \bibliographystyle{ieee_fullname}
    \bibliography{egbib}
}

\onecolumn
\section*{\hfil {\LARGE Appendix}\hfil}
\vspace{50pt}


\renewcommand\thesection{\Alph{section}}
\renewcommand*{\theHsection}{appedix.\thesection}
\setcounter{section}{0}
\setcounter{figure}{6}
\setcounter{table}{6}
\setcounter{equation}{4}

This appendix provides more qualitative results (\cref{sec:appendix:qualitative}), 
several additional experiments (\cref{sec:appendix:exps}) on the components of motion latent diffusion (MLD) models, 
evaluations of inference time (\cref{sec:appendix:inferencetime}),
visualization of latent space (\cref{sec:appendix:tsne}),
evaluations on hyperparameters (\cref{sec:appendix:hyperparameters}), 
user study (\cref{sec:appendix:userstudy}), 
details of motion representations (\cref{sec:appendix:motionRepre}), 
implementation details of MLD models (\cref{appedix:method:detail}) and metric definitions (\cref{appedix:metrics:details}).\\



\myparagraph{Video.} We have provided supplemental videos in \href{https://chenxin.tech/mld}{Project Page}. In these supplemental videos, we show 1) comparisons of text-based motion generation, 2) comparisons of action-conditional motion generation, and 3) more samples of unconditional generation. We suggest the reader watch this video for dynamic motion results.\\

\myparagraph{Code} is available on \href{https://github.com/chenfengye/motion-latent-diffusion}{GitHub Page}. We provide the process of the training and evaluation of MLD models, the pre-trained model files, the demo script, and example results.

\section{Qualitative Results}
\begin{figure}[h]
    \centering
    \includegraphics[width=\linewidth]{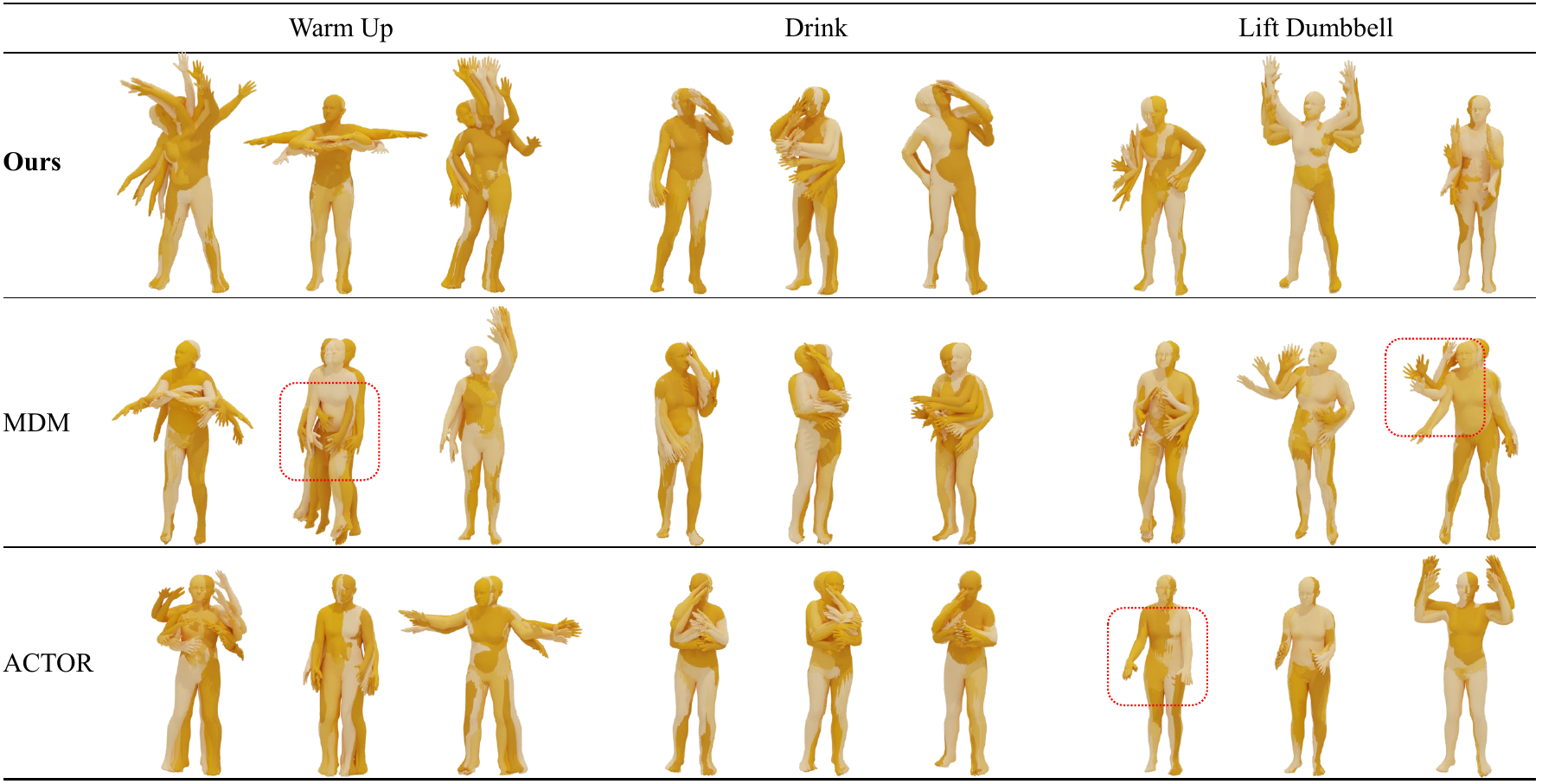}
    \caption{The comparison of the state-of-the-art methods on action-conditional motion synthesis task. All provided methods are under the same training and inference setting on HumanAct12 dataset~\cite{guo2020action2motion}. We generate three motions for each action label. The results demonstrate that our generations correspond better to the action label and have richer diversity.}
    \label{fig:appendix:action}
\end{figure}

\label{sec:appendix:qualitative}
\begin{figure}[h]
    \centering
    \includegraphics[width=0.955\linewidth]{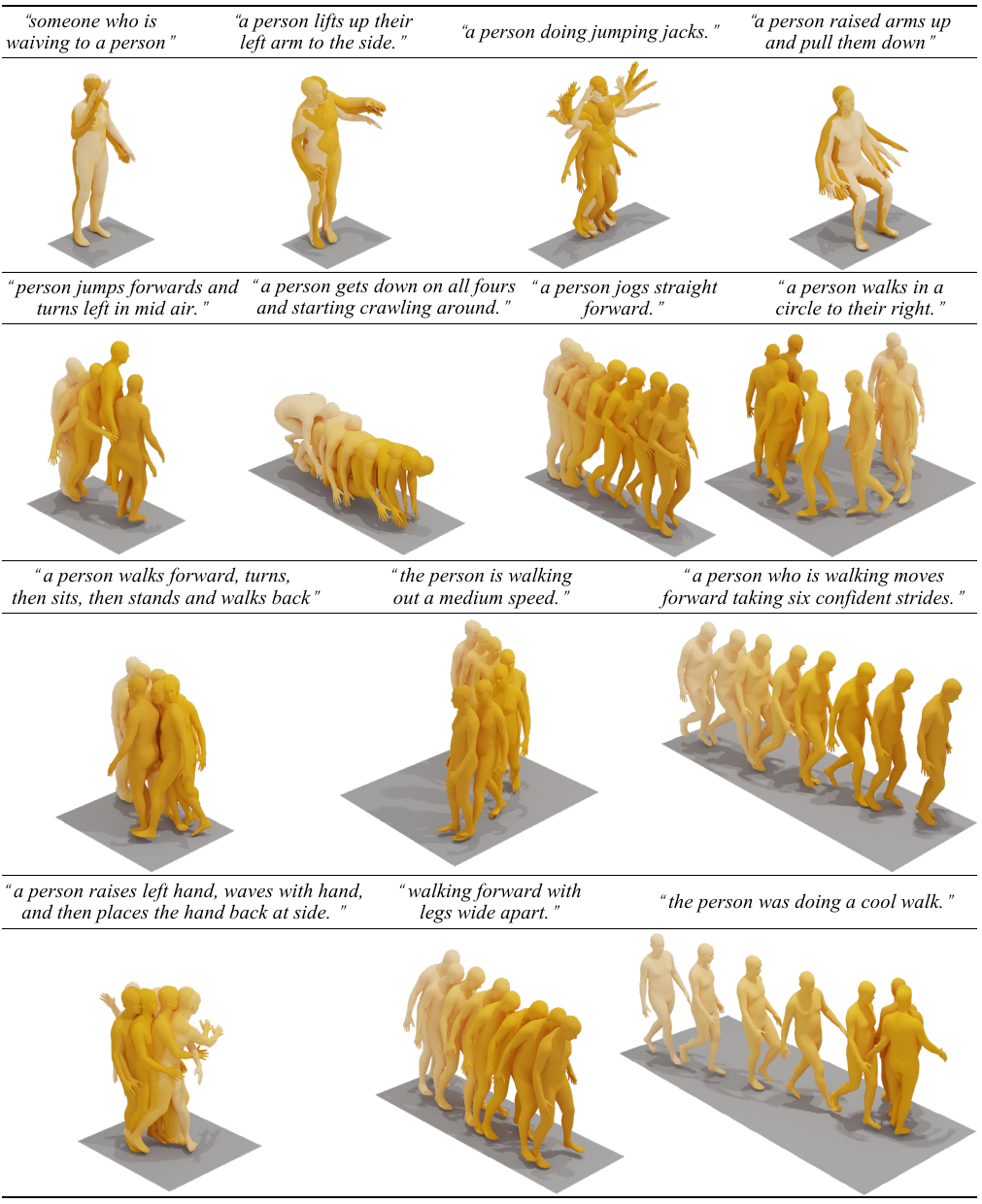}
    \caption{More samples from our best model for text-to-motion synthesis, \textit{MLD-1}, which was trained on the HumanML3D dataset. Samples generated with text prompts of the test set. We recommend the supplemental video to see these motion results. }
    \label{fig:appendix:more_rs}
\end{figure}

  
        

\newpage
\section{Additional Experiments}
\label{sec:appendix:exps}
We conduct several experiments to continue the evaluations of MLD models. We first study the influence of language models $\tau_\theta^w$ and the shape of text embedding on motion generations. Then, we evaluate the effectiveness of long skip connections for motion diffusion models. We finally study the importance of regularization on motion latent space.  


\subsection{Evaluation of Language Models $\tau_\theta^w$ }
We experiment with different language models, CLIP~\cite{radford2021learning} and BERT~\cite{devlin2018bert}. 
Inspired by Stable Diffusion~\cite{stable_diffusion}, we leverage the hidden state of CLIP to generate word-wise tokens and explore its effects.
The comparisons are listed in \cref{tab:appendix:language}. CLIP is more suited to our task compared to BERT, and the word-wise text tokens are competitive with the single token, however, lower the computation efficiency of diffusion models.
Therefore, we choose CLIP and a single text token for our models.

\begin{table}[h]
\centering
\resizebox{0.8\columnwidth}{!}{%
\begin{tabular}{@{}lccccccc@{}}
\toprule
\multirow{2}{*}{Models} & Text Encoder& Embeddings & \multicolumn{1}{c}{R Precision} & \multicolumn{1}{c}{\multirow{2}{*}{FID$\downarrow$}} &  \multirow{2}{*}{MM Dist$\downarrow$} & \multirow{2}{*}{Diversity$\rightarrow$} & \multirow{2}{*}{MModality$\uparrow$} \\
                          & $\tau_\theta^w$ &Shape & \multicolumn{1}{c}{Top 3$\uparrow$}       & \multicolumn{1}{c}{}                     &                          &                            &     \\ \toprule
Real & - & - &
  $0.797^{\pm.002}$ &
  $0.002^{\pm.000}$ &
  $2.974^{\pm.008}$ &
  $9.503^{\pm.065}$ &
  \multicolumn{1}{c}{-}
  \\ \midrule
MLD-1 & BERT~\cite{devlin2018bert}& $1\times256$ 
&$0.725^{\pm.002}$&	$0.553^{\pm.020}$&	$3.530^{\pm.011}$&	$\boldsymbol{9.697}^{\pm.080}$&	$\boldsymbol{3.360}^{\pm.118}$
  \\
MLD-1 & CLIP~\cite{radford2021learning}   & $1\times256$
& $\boldsymbol{0.769}^{\pm.002}$ &	$0.446^{\pm.011}$&	$\boldsymbol{3.227}^{\pm.010}$&	$9.772^{\pm.071}$&	$2.413^{\pm.072}$\\
MLD-1 & CLIP~\cite{radford2021learning} & $77\times256$& $0.737^{\pm.002}$	& $\boldsymbol{0.422}^{\pm.012}$	& $3.436^{\pm.010}$ &	$9.840^{\pm.082}$ &	$2.799^{\pm.107}$   \\

\bottomrule
\end{tabular}%
}
\caption{Quantitative comparison of the employed language models. Here we set batch size to 500 and only change the text encoder $\tau_\theta^w$. }
\label{tab:appendix:language}
\end{table}

\subsection{Effectiveness of Long Skip Connection}
\begin{figure}[h]
    \centering
    \includegraphics[width=0.7\linewidth]{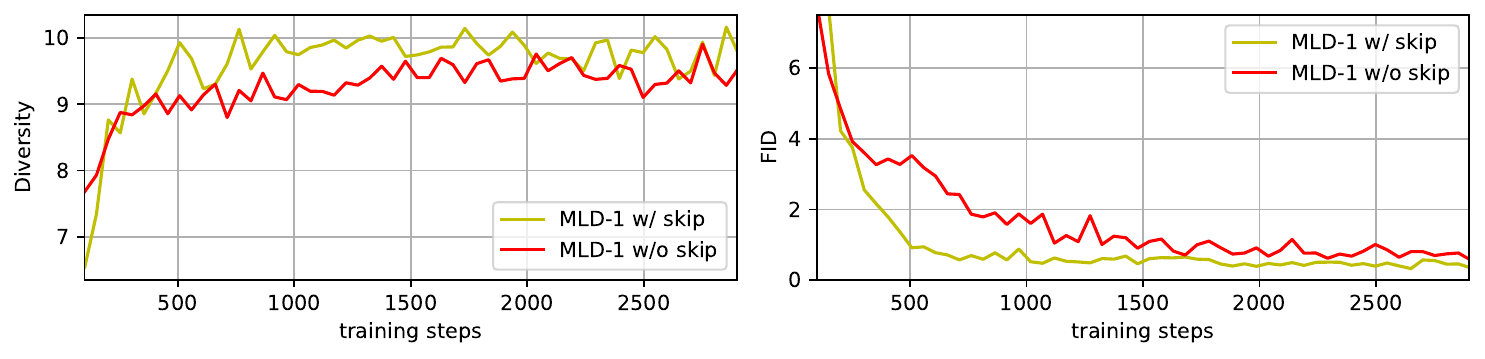}
    \vspace{-10pt}
    \caption{The evaluation on long skip connection on diffusion training stage. Two sub-figures are under the same training process and evaluated on the test set of HumanML3D. Training steps indicate the epoch number.}
    \label{fig:appendix:skip}
\end{figure}

We have demonstrated the effectiveness of skip connection, especially on diffusion models in \cref{tab:ablation:diffusion}. Here we analyze its influence on the training of diffusion stage. As shown in \cref{fig:appendix:skip}, the model with long skip connection not only achieves higher performance but also provides faster convergence compared to the other one. The results suggest leveraging long skip connections for iterative motion diffusion models.

\subsection{Diffusion on Autoencoder or VAE}
We study the importance of regularization on motion latent space. The regularized latent space provides stronger generation ability and supports the latent diffusion models as demonstrated:

\begin{table}[H]
\centering
\resizebox{0.65\columnwidth}{!}{%
\begin{tabular}{@{}lccccccl@{}}
\toprule
\multirow{2}{*}{Method} & \multicolumn{3}{c}{Reconstruction}                            &  \multicolumn{2}{c}{Generation}        \\ \cmidrule(lr){2-4} \cmidrule(l){5-6} 
                        & MPJPE $\downarrow$ & PAMPJPE$\downarrow$ & ACCL$\downarrow$ &   FID$\downarrow$& DIV$\rightarrow$\\ \midrule

Autoencoder & {38.5} &  {28.2} & 5.8 & 0.156 & 9.628
\\
VAE & $\boldsymbol{14.7}$ & $ \boldsymbol{8.9}$ & $\boldsymbol{5.1}$ & $\boldsymbol{0.017}\hidden{^{\pm.000}}$ &  $\boldsymbol{9.554}\hidden{^{\pm.062}}$
\\ 
\bottomrule
\toprule

\multirow{2}{*}{Method} & \multicolumn{1}{c}{R Precision} & \multicolumn{1}{c}{\multirow{2}{*}{FID$\downarrow$}} & \multirow{2}{*}{MM Dist$\downarrow$} & \multirow{2}{*}{Diversity$\rightarrow$} & \multirow{2}{*}{MModality$\uparrow$} \\
                          & \multicolumn{1}{c}{Top 3$\uparrow$}       & \multicolumn{1}{c}{}                     &                          &                            &     \\ \toprule
MLD w/ Autoencoder & $0.581^{\pm.003}$ &	$5.033^{\pm.061}$ &	$4.600^{\pm.018}$ &	$7.953^{\pm.083}$ &	$\boldsymbol{3.754}^{\pm.111}$
 \\
MLD w/ VAE & 
$\boldsymbol{0.772}^{\pm.002}$ &	$\boldsymbol{0.473}^{\pm.013}$ &	$\boldsymbol{3.196}^{\pm.010}$ &	$\boldsymbol{9.724}^{\pm.082}$ &	$2.413^{\pm.079}$

\\ \bottomrule
\end{tabular}%
}
\caption{Evaluation of autoencoder (without  Kullback-Leibler regularization) and VAE model on motion generations.}
\label{tab:supp:ablation}
\end{table}

\subsection{Prediction of Denoising}
\label{sec:appendix:predictdenoising}

We compare predicting the denoised latent vector $z_0$ directly instead of $\epsilon$ in the denoising process. 
\cref{tab:tm:abl:humanml3d} shows that the latter performs better, which verifies the proposal from DDPM ~\cite{ho2020denoising}.

\begin{table}[h]
\centering
\resizebox{0.8\columnwidth}{!}{%
\begin{tabular}{@{}lccccccc@{}}
\toprule
\multirow{2}{*}{Methods} & \multicolumn{3}{c}{R Precision $\uparrow$}& \multicolumn{1}{c}{\multirow{2}{*}{FID$\downarrow$}} & \multirow{2}{*}{MM Dist$\downarrow$}& \multirow{2}{*}{Diversity$\rightarrow$} & \multirow{2}{*}{MModality$\uparrow$}\\ \cmidrule(lr){2-4} & \multicolumn{1}{c}{Top 1} & \multicolumn{1}{c}{Top 2} & \multicolumn{1}{c}{Top 3} & \multicolumn{1}{c}{} &&&\\ \midrule
MLD-1 ($z_0$) &
  $0.447^{\pm.002}$ &
  $0.633^{\pm.002}$ &
  $0.734^{\pm.002}$ &
  $0.513^{\pm.011}$ &
  $3.384^{\pm.008}$ &
  $9.181^{\pm.065}$ &
  $0.735^{\pm.055}$ 
   \\ 
MLD-1 ($\epsilon$) &
  $\boldsymbol{0.481}^{\pm.003}$ &
  $\boldsymbol{0.673}^{\pm.003}$ &
  $\boldsymbol{0.772}^{\pm.002}$ &
  $\boldsymbol{0.473}^{\pm.013}$ &
  $\boldsymbol{3.196}^{\pm.010}$ &
  $\boldsymbol{9.724}^{\pm.082}$ &
  $\boldsymbol{2.413}^{\pm.079}$ 
   \\   
\bottomrule
\end{tabular}%
}
\caption{Comparison of text-to-motion. ($cf.$ Tab. 1 for details.)}
\label{tab:tm:abl:humanml3d}
\end{table}

\section{Inference time}
\label{sec:appendix:inferencetime}
We provide a detailed ablation study with DDIM below.
In \cref{tab:tm:abl:scheduler}, 
MLD reduces the computational cost of diffusion models, which is the main reason for faster inference.
The iterations of diffusion further widen the gap in computational cost.
Please note the bad FID of MDM with DDIM is mentioned in their GitHub issues \#76.
\begin{table}[h]
\centering
\resizebox{\columnwidth}{!}
{%
\begin{tabular}{@{}lrrrrrrrrlccccc@{}}
\toprule
\multirow{4}{*}{Methods} & \multicolumn{4}{c}{Total Inference Time (s) $\downarrow$} & \multicolumn{4}{c}{FLOPs (G) $\downarrow $} & \multirow{4}{*}{Parameter} & \multicolumn{4}{c}{FID $\downarrow $} \\
\cmidrule(lr){2-5} \cmidrule(lr){6-9} \cmidrule(lr){11-14}&\multicolumn{3}{c}{DDIM}& DDPM&\multicolumn{3}{c}{DDIM}& DDPM&&\multicolumn{3}{c}{DDIM}&DDPM\\
\cmidrule(lr){2-4} \cmidrule(lr){5-5} \cmidrule(lr){6-8} \cmidrule(lr){9-9} \cmidrule(lr){11-13} \cmidrule(lr){14-14}
& 50 & 100 & 200 & 1000& 50 & 100 & 200 & 1000&&50&100 & 200&1000\\

 \toprule
 MDM
&225.28 & 456.70 & 911.36 & \textcolor{black}{4546.23}
&597.97&1195.94&2391.89&11959.44&
$x \in \mathbb{R}^{196\times512}$
& 7.334&5.990&5.936&0.544\\ 

MLD
&\textbf{10.24} & \textbf{16.38} & \textbf{28.67 }& \textbf{148.97}
&\textbf{29.86}&\textbf{33.12}&\textbf{39.61}&\textbf{91.60}& 
$z \in \mathbb{R}^{1\times256}$ 
&0.473&0.426&0.432&\textcolor{black}{0.568}\\ 

\bottomrule
\end{tabular}%
}
\caption{Evaluation of inference time costs on text-to-motion: we evaluate the total inference time to generate 2048 motion clips with different diffusion schedules, floating point operations (FLOPs) counted by THOP library, the size of diffusion input, and FID.}
\label{tab:tm:abl:scheduler}
\end{table}

\section{Latent space visualization}
\label{sec:appendix:tsne}

\begin{figure}[htb]
    \centering
    \includegraphics[width=1.0\linewidth]{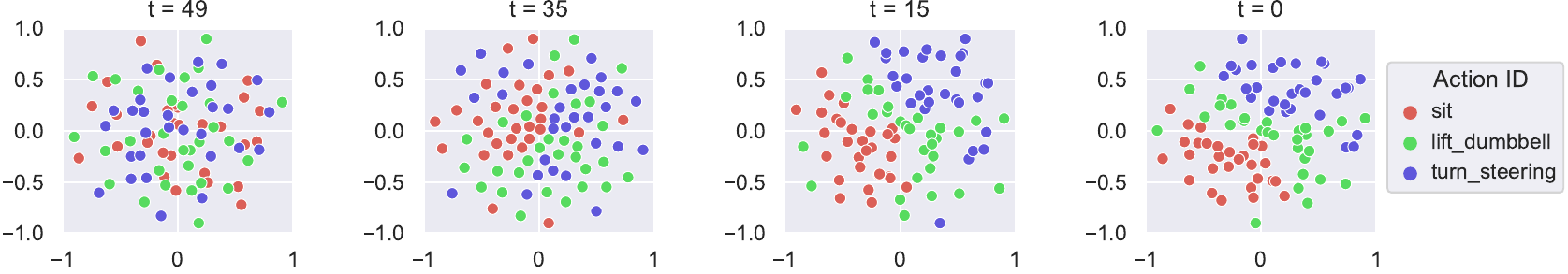}
    \caption{Visualization of the t-SNE results on evolved latent codes $\hat{z}_t$ during the reverse diffusion process (inference) on action-to-motion task.
    $t$ is the diffusion step but ordered in the forward diffusion trajectory.
    $\hat{z}_{t=49}$ is the initial random noise. $\hat{z}_{t=0}$ is our prediction.
    We sample 30 motions for each action label.}
    \label{fig:tsne}
\end{figure}

We provide the visualizations of the t-SNE results on the latent space in  \cref{fig:tsne} to demonstrate how latent space evolves during the diffusion process with different actions.
From left to right, it shows the evolved latent codes during the inference of diffusion models.

\section{Evaluation of Hyperparameters}
\label{sec:appendix:hyperparameters}

Here, we present two different experiments of text-to-motion on HumanML3D~\cite{Guo_2022_CVPR_humanml3d}. The first experiment is to change the dropout $p$ and scale in classifier-free diffusion guidance~\cite{ho2022classifier}. In \cref{tab:appendix:guidance}, we find that by changing dropout $p$ from $0.1$ to $0.25$, the text correspondences (R Precision) become worse but the motion quality (FID) gets better. It is the same as changing scale $s$ range from $7.5$ to $2.5$. Besides, some settings like $(0.25,7.5)$ achieve the best FID of 0.229, but we still suggest $(0.1,7.5)$ as dropout and scale $(p,s)$ for MLD models (\cref{experiments}) overall metrics.

Next, in \cref{tab:appendix:batchsize}, we experiment with batch sizes of 32, 64, 128, 256 and 512 under 8 Tesla V100 each with 32 GPU memory. We set it to 64 in our other experiments.
\label{appendix:guidance}



\begin{table}[H]
\centering
\resizebox{0.8\columnwidth}{!}{%
\begin{tabular}{@{}lccccccc@{}}
\toprule
\multirow{2}{*}{Models}  & \multicolumn{2}{c}{Classifier-free} & \multicolumn{1}{c}{R Precision} & \multicolumn{1}{c}{\multirow{2}{*}{FID$\downarrow$}} & \multirow{2}{*}{MM Dist$\downarrow$} & \multirow{2}{*}{Diversity$\rightarrow$} & \multirow{2}{*}{MModality$\uparrow$} \\\cmidrule(l){2-3}
                          & Dropout & Scale & \multicolumn{1}{c}{Top 3$\uparrow$}       & \multicolumn{1}{c}{}                     &                          &                            &     \\ \toprule
Real & - &- &
  $0.797^{\pm.002}$ &
  $0.002^{\pm.000}$ &
  $2.974^{\pm.008}$ &
  $9.503^{\pm.065}$ &
  \multicolumn{1}{c}{-}
  \\ \midrule
MLD-1& $p=0.05$ &  $s=7.5$ &
  $0.766^{\pm.002}$	&$0.574^{\pm.013}$	&$3.237^{\pm.007}$&	$9.664^{\pm.069}$&	$2.433^{\pm.074}$ \\
MLD-1 &$p=0.10$  & $s=7.5$    & $\boldsymbol{0.772}^{\pm.002}$ &	$0.473^{\pm.013}$ &	$\boldsymbol{3.196}^{\pm.010}$ &	$9.724^{\pm.082}$ &	$2.413^{\pm.079}$\\
MLD-1 &$p=0.15$ & $s=7.5$ & $0.765^{\pm.002}$&	$0.311^{\pm.009}$&	$3.209^{\pm.007}$&	$9.649^{\pm.065}$&	$2.525^{\pm.070}$  \\
MLD-1 &$p=0.20$  & $s=7.5$ &$0.761^{\pm.002}$& $0.279^{\pm.011}$ &	$3.243^{\pm.009}$ &	$\boldsymbol{9.632}^{\pm.082}$ &	$2.651^{\pm.080}$\\
MLD-1 &$p=0.25$  & $s=7.5$ & $0.757^{\pm.002}$& $\boldsymbol{0.229}^{\pm.010}$&	$3.260^{\pm.008}$&	$9.649^{\pm.069}$&	$\boldsymbol{2.685}^{\pm.084}$\\
MLD-1 &$p=0.30$  & $s=7.5$ & $0.759^{\pm.002}$&	$0.289^{\pm.010}$&	$3.249^{\pm.008}$&	$9.670^{\pm.073}$&	$2.650^{\pm.082}$
\\
\midrule
MLD-1 &$p=0.10$  & $s=1.5$ &$0.648^{\pm.002}$&	$0.401^{\pm.019}$&	$3.857^{\pm.009}$&	$9.263^{\pm.056}$&$\boldsymbol{3.914}^{\pm.115}$\\
MLD-1 &$p=0.10$  & $s=2.5$ & $0.720^{\pm.002}$	&$\boldsymbol{0.350}^{\pm.013}$&	$3.441^{\pm.010}$	&$\boldsymbol{9.549}^{\pm.058}$&	$3.201^{\pm.098}$\\
MLD-1 &$p=0.10$  & $s=3.5$ &$0.745^{\pm.002}$&	$0.358^{\pm.011}$&	$3.299^{\pm.009}$&	$9.639^{\pm.065}$&	$2.890^{\pm.087}$\\
MLD-1 &$p=0.10$  & $s=4.5$ & $0.758^{\pm.002}$ &	$0.375^{\pm.011}$&	$3.232^{\pm.009}$&	$9.676^{\pm.069}$&	$2.701^{\pm.078}$ \\
MLD-1 &$p=0.10$  & $s=5.5$ & $0.764^{\pm.002} $	&$0.396^{\pm.011}$&	$3.202^{\pm.010}$&	$9.681^{\pm.072}$&	$2.577^{\pm.076}$  \\
MLD-1 &$p=0.10$  & $s=6.5$ & $0.767^{\pm.002}$ &$0.424^{\pm.011}$&	$\boldsymbol{3.191}^{\pm.009}$&	$9.658^{\pm.072}$& 	$2.498^{\pm.074}$  \\
MLD-1 &$p=0.10$  & $s=7.5$ & $\boldsymbol{0.772}^{\pm.002}$ &	$0.473^{\pm.013}$ &	$3.196^{\pm.010}$ &	$9.724^{\pm.082}$ &	$2.413^{\pm.079}$\\
MLD-1 &$p=0.10$  & $s=8.5$ & $0.768^{\pm.002}$ &	$0.504^{\pm.012}$ &	$3.207^{\pm.009}$ &	$9.604^{\pm.073}$ &	$2.413^{\pm.072}$  \\
MLD-1 &$p=0.10$  & $s=9.5$ & $0.766^{\pm.001}$&	$0.555^{\pm.012}$&	$3.227^{\pm.010}$&	$9.567^{\pm.072}$&	$2.394^{\pm.069}$ \\
\bottomrule
\end{tabular}%
}
\vspace{-6pt}
\caption{\textbf{Classifier-free Diffusion Guidance:} We study the influence of its hyperparameters, dropout $p$ and scale $s$ on text-to-motion.}
\label{tab:appendix:guidance}
\vspace{-6pt}
\end{table}



\begin{table}[H]
\centering
\vspace{-12pt}
\resizebox{0.75\columnwidth}{!}{%
\begin{tabular}{@{}lcccccc@{}}
\toprule
\multirow{2}{*}{Models}  & \multicolumn{1}{c}{\multirow{2}{*}{Batch Size}} & \multicolumn{1}{c}{R Precision} & \multicolumn{1}{c}{\multirow{2}{*}{FID$\downarrow$}} & \multirow{2}{*}{MM Dist$\downarrow$} & \multirow{2}{*}{Diversity$\rightarrow$} & \multirow{2}{*}{MModality$\uparrow$} \\
                          &  & \multicolumn{1}{c}{Top 3$\uparrow$}       & \multicolumn{1}{c}{}                     &                          &                            &     \\ \toprule
Real &-&
  $0.797^{\pm.002}$ &
  $0.002^{\pm.000}$ &
  $2.974^{\pm.008}$ &
  $9.503^{\pm.065}$ &
  \multicolumn{1}{c}{-}
  \\ \midrule
MLD-1 & 32 & $0.761^{\pm.003}$&	$0.445^{\pm.012}$&	$3.243^{\pm.010}$&	$9.751^{\pm.086}$&	$\boldsymbol{2.581}^{\pm.070}$
   \\
MLD-1 & 64  & $\boldsymbol{0.772}^{\pm.002}$ &	$0.473^{\pm.013}$ &	$\boldsymbol{3.196}^{\pm.010}$ &	$9.724^{\pm.082}$ &	$2.413^{\pm.079}$\\
MLD-1 & 128 & $0.771^{\pm.002}$&	$\boldsymbol{0.421}^{\pm.013}$&	$3.187^{\pm.008}$&	$\boldsymbol{9.691}^{\pm.080}$&	$2.370^{\pm.078}$ \\
MLD-1 & 256 & 
$0.770^{\pm.002}$&	$0.423^{\pm.010}$&	$3.211^{\pm.007}$&	$9.800^{\pm.070}$&	$2.401^{\pm.074}$
\\
MLD-1 & 512 & $0.769^{\pm.002}$ &	$0.446^{\pm.011}$&	$3.227^{\pm.010}$&	$9.772^{\pm.071}$&	$2.413^{\pm.072}$  \\

\bottomrule
\end{tabular}%
}
\caption{\textbf{Batch Size:} We explore the evaluation of the batch size. We find the results are close and suggest 64 and 128 in this task.}
\label{tab:appendix:batchsize}
\end{table}


\section{User Study}
\label{sec:appendix:userstudy}
\begin{figure}[h]
    \centering
    \includegraphics[width=0.9\linewidth]{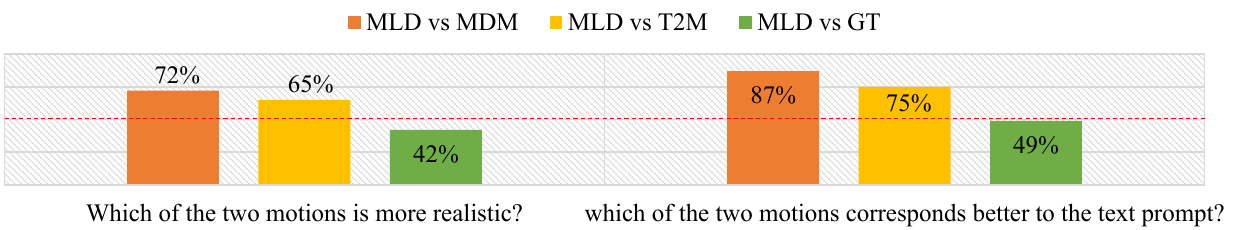}
    \caption{\textbf{User Study:} Each bar indicates the preference rate of MLD over other methods. The red line indicates the 50\%. Please refer to the supplemental video for the comparisons of dynamic motion results.}
    \label{fig:appendix:user}
\end{figure}

For the pairwise comparisons of the user study presented in \cref{fig:appendix:user}, we use the force-choice paradigm to ask ``Which of the two motions is more realistic?'' and ``which of the two motions corresponds better to the text prompt?''. The provided motions are generated from 30 text descriptions, which are randomly generated from the test set of HumanML3D\cite{Guo_2022_CVPR_humanml3d} dataset. We invite 20 users and provide three comparisons, ours and MDM~\cite{mdm2022human}, ours and T2M~\cite{Guo_2022_CVPR_t2m}, ours and real motions from the dataset. 
Our MLD was preferred over the other state-of-the-art methods and even competitive to the ground troth motions.

\section{Motion Representations}
\label{sec:appendix:motionRepre}
Four relevant motion representations are summarized:

 \textbf{HumanML3D Format}~\cite{Guo_2022_CVPR_humanml3d} proposes a motion representation $x^{1:L}$ inspired by motion features in character control~\cite{starke2019neural, 2021-TOG-AMP, starke2022deepphase}. This redundant representation is quite suited to neural models, particularly variational autoencoders. Specifically, the $i$-th pose $x^i$ is defined by a tuple of root angular velocity $\dot{r}^a \in \mathbb{R}$ along Y-axis, root linear velocities $(\dot{r}^x, \dot{r}^z \in \mathbb{R})$ on XZ-plane, root height $r^y \in \mathbb{R}$, local joints positions $\mathbf{j}^p\in\mathbb{R}^{3N_j}$, velocities $\mathbf{j}^v\in\mathbb{R}^{3N_j}$ and rotations $\mathbf{j}^r\in\mathbb{R}^{6N_j}$ in root space, and binary foot-ground contact features $\mathbf{c}^f \in \mathbb{R}^4$ by thresholding the heel and toe joint velocities, where $N_j$ denotes the joint number, giving:
\begin{align}
 x^i = \{\dot{r}^a, \dot{r}^x, \dot{r}^z, r^y, \mathbf{j}^p, \mathbf{j}^v, \mathbf{j}^r, \mathbf{c}^f\}.
\end{align}
%
%

\textbf{SMPL-based Format}~\cite{SMPL2015}. The most popular parametric human model, SMPL~\cite{SMPL2015} and its variants~\cite{MANO:SIGGRAPHASIA:2017, SMPLX2019} propose motion parameters $\theta$ and shape parameters $\beta$. $\theta \in \mathbb{R}^{3\times23+3}$ is rotation vectors for 23 joints and a root, and $\beta$ are the weights for linear blended shapes.
This representation is popular in markerless motion capture~\cite{he2021challencap,chen2021sportscap,VIBE_CVPR2020}.
By involving the global translation $r$, the representation is formulated as:
\begin{align}
 x^i = \{r, \theta, \beta\}.
\end{align}


\textbf{MMM Format}~\cite{terlemez2014master}. Master Motor Map (MMM) representations propose joints angle parameters by adopting a uniform skeleton structure with 50 DoFs. And most recent methods ~\cite{ahuja2019language2pose,ghosh2021synthesis,petrovich22temos} on text-to-motion task followed preprocess procedure in ~\cite{holden2016deep} which transform joint rotation angles to $J=21$ joints XYZ coordinates, giving $p_m\in \mathbb{R}^{3J}$,  and global trajectory $t_{root}$ for the root joint. 
The preprocessed representation can be formulated as 
\begin{align}
 x^i = \{p_m, t_{root}\}.
\end{align}


\textbf{Latent Format}~\cite{SMPL2015}. 
Latent representations are widely used in neural models~\cite{petrovich21actor, petrovich22temos,guo2020action2motion, chen2022learning}.
We recognize it as motion representation in latent space.
By leveraging VAE models, latent vectors can represent plausible motions as:
\begin{align}
 \hat{x}^{1:L} = \mathcal{D}(z), z = \mathcal{E}(x^{1:L})
\end{align}

\section{Details on Motion Latent Diffusion Models}
\label{appedix:method:detail}
\subsection{Details Information on Variational Autoencoder Models}
\label{sec:appendix:loss}
\jb{We take HumanML3D~\cite{Guo_2022_CVPR_humanml3d} and its motion representation (\cref{sec:appendix:motionRepre}) as an example here to explain our loss details of Variational Autoencoder Models $\mathcal{V}$.}
%
%
The motion $x^{1:L}$ includes joint features and is supervised with data term by mean squared error:
\begin{align}
\mathcal{L}_{data} = \left\| x^{1:L} - \mathcal{D}(\mathcal{E}(x^{1:L})) \right\|^2.
\end{align}
To regularize latent space as a standard variational autoencoder~\cite{kingma2013auto}, we employ a Kullback-Leibler term between $q(z|x^{1:L}) = \mathcal{N}\left(z ; \mathcal{E}_\mu, \mathcal{E}_{\sigma^2}\right)$ and a standard Gaussion distribution $\mathcal{N}\left(z;0,1\right)$.
The full training loss of the VAE model $\mathcal{V}$ follows:
\begin{align}
\mathcal{L}_{\mathcal{V}} = \mathcal{L}_{data}(x^{1:L}, \mathcal{D}(\mathcal{E}(x^{1:L}))) + \lambda_{reg}\mathcal{L}_{reg}(x^{1:L};\mathcal{E},\mathcal{D}),
\end{align}
where $\lambda_{reg}$ is a low weight to control the regularization. 
The KIT~\cite{Plappert2016kit}, HumanAct12~\cite{guo2020action2motion} and UESTC~\cite{ji2018large} dataset processed by \cite{petrovich22temos,petrovich21actor} also supports SMPL-based~\cite{SMPL2015} motion representation.
Here we list the loss terms for this representation.
The data term formulates as followed:
\begin{align}
    \label{eq:vae:humanml}
    \mathcal{L}_{data} = \sum_{i=1}^L \left\|{r}^i  -\hat{{r}}^i \right\|_2 + \sum_{i=1}^L \left\|{\theta}^i  -\hat{{\theta}}^i \right\|_2 +  \left\|{\beta}  -\hat{{\beta}} \right\|_2.
\end{align}
Here the motion is $x^{1:L} = \{{r}^i, {\theta}^i, {\beta}\}^{L}_{i=1}$, which includes global translation ${r}^i$, pose parameter ${\theta}^i$ and shape parameter ${\beta}$ of the $i$-th frame. 
To enhance the full-body supervision, the reconstruction term on the SMPL vertices follows:
\begin{align}
    \mathcal{L}_{\mathrm{mesh}} = \sum_{i=1}^L \left\| {V}_i - M(\hat{{r}}^i, \hat{{\theta}}^i,\hat{{\beta}}^i) \right\|^2,
\end{align}
where the body reconstruction function $M(\cdot)$ is from the differentiable SMPL layer, while the vertices ${V}_i$ are calculated with the ground truth motion parameters using the same layer.
The reconstruction loss builds global supervision on almost all predicted parameters $ \{{r}_t, {\theta}_t, {\beta}\} $ and shows a reliable supervision~\cite{petrovich21actor} for motion generation.
The full objective on SMPL-based motion representation reads:
\begin{align}
    \label{eq:vae:smpl}
    \mathcal{L}_{\mathcal{V}} = \mathcal{L}_{data} + \lambda_{mesh}\mathcal{L}_{mesh} + \lambda_{reg}\mathcal{L}_{reg}.
\end{align}
where $\lambda_{mesh}$ is the weight to enhance the supervision on the full-body vertices. Besides, the regularization term is the same as the Kullback-Leibler term in \cref{eq:vae:humanml}.
In practice, the shape parameters, as part of global motion features, increase the complexity of motion generation and influence joint positions. 
We finally utilize the objective of ~\cref{eq:vae:humanml} to train our text-based models and \cref{eq:vae:smpl} to train action-based models in comparisons and evaluations.

\subsection{Network Architectures}
\begin{figure}[h]
    \centering
    \includegraphics[width=0.85\linewidth]{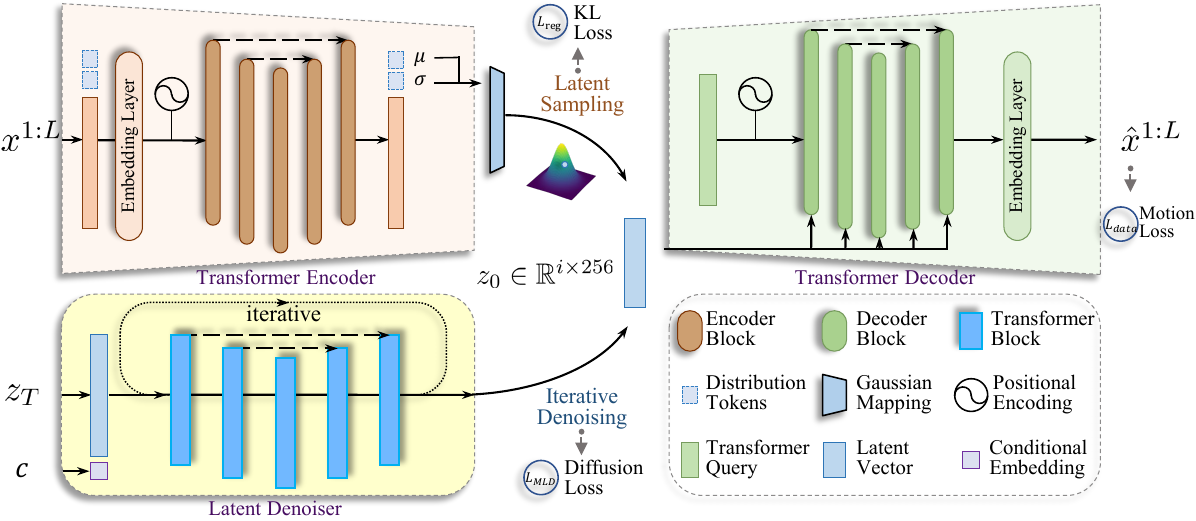}
    \caption{Network architecture of our conditional MLD. We explain each component in its right bottom part and the loss terms in \cref{sec:appendix:loss}.  }
    \label{fig:appendix:net}
\end{figure}

The details of network architecture are shown as \cref{fig:appendix:net}, our MLD comprises three main components, motion encoder $\mathcal{E}$, motion decoder $\mathcal{D}$ and latent denoiser $\epsilon_\theta$. Please refer to the following figure and \cref{tab:appendix:details} for more details.

\subsection{Implementation Details}
For the experiments on text-to-motion, action-to-motion, and unconditional motion synthesis, we implement MLDs with various latent shapes as follows. Specifically, MLD-7 works best in evaluating VAE models (\cref{tab:mr:ablation}), and MLD-1 wins these generation tasks (\cref{tab:tm:comp:humanml3d,,tab:tm:comp:kit,,tab:comp:action,,tab:ablation:uncon}). In other words, MLD-7 wins the first training stage for the VAE part, while MLD-1 wins the second for the diffusion part. We thought MLD-7 should perform better than MLD-1 in several tasks, but the results differ. The main reason for this downgrade of a larger latent size, we believe, is the small amount of training data. HumanML3D only includes 15k motion sequences, much smaller than billions of images in image generation. MLD-7 could work better when the motion data amount reaches the million level.

\begin{table}[H]
\centering
\resizebox{0.7\columnwidth}{!}{%
\begin{tabular}{@{}lccccc@{}}
\toprule
\multirow{2}{*}{} &  MLD-1 & MLD-2 & MLD-5 & MLD-7 & MLD-10
 \\ \toprule
$z$-shape & $1\times 256$& $2\times 256$& $5\times 256$& $7\times 256$& $10\times 256$
\\
Training Diffusion steps & 1000 &1000 &1000 &1000 & 1000 
\\
Inference Diffusion steps  & 50 &50 &50 &50 & 50 
\\
Noise Schedule & scaled linear & scaled linear & scaled linear & scaled linear & scaled linear
\\
Denoiser Heads Number & 4 & 4& 4& 4& 4
\\
Denoiser Transformer Layers & 9 &9 & 9 & 9 & 9 
\\
Conditioning & concat& concat& concat& concat& concat
\\
\midrule
Embedding Dimension & 256 & 256 & 256 & 256 & 256
\\
VAE Heads Number  & 4 & 4& 4& 4& 4
\\
VAE Transformer Layers & 9 & 9 & 9 & 9 & 9 
\\\midrule
Model Size (w/o clip) & 26.9M&26.9M&26.9M&26.9M&26.9M
\\
Diffusino Batch Size & 64 &64 &64 &64 &64
\\
Diffusion Epochs & 2000& 2200& 2400& 2600& 2800
\\
VAE Batch Size & 128 &128 &128 &128 &128
\\
VAE Epochs & 4000& 4500& 5000& 5500& 6000
\\
Learning Rate & 1e-4& 1e-4& 1e-4& 1e-4& 1e-4
\\
\bottomrule
\end{tabular}%
}
\caption{Hyperparameters for the conditional MLDs in experiments. We train these models on 8 Tesla V100. The smaller latent shape lowers the computational requirements and provides faster inference. }
\label{tab:appendix:details}
\end{table}

\section{Metric Definitions}
\label{appedix:metrics:details}

We provide more details of evaluation metrics in \cref{sec:comp:metric} as follows.

\textbf{Motion Quality}. \jb{
Frechet Inception Distance (FID) is our principal metric to evaluate the distribution similarity between generated and real motions, calculated with the suitable feature extractor~\cite{guo2020action2motion,petrovich21actor,Guo_2022_CVPR_t2m} for each dataset. 
}
Besides, to evaluate the motion reconstruction error of VAEs, we use popular metrics in motion capture~\cite{VIBE_CVPR2020, chen2021sportscap, vonMarcard2018}, 
MPJPE, and PAMPJPE~\cite{gower1975generalized} for global and local errors in millimeter, 
Acceleration Error (ACCL) for the quality on temporal.

\textbf{Generation Diversity}. \jb{Following ~\cite{guo2020action2motion,chuan2022tm2t}, we use Diversity (DIV) and MultiModality (MM) to measure the motion variance across the whole set and the generated motion diversity within each text input separately. Here we take the text-to-motion task as an example to explain the calculation steps and for other tasks the operations are similar. To evaluate Diversity, all generated motions are randomly sampled to two subsets of the same size $X_d$ with motion feature vectors $\{x_1,..,x_{X_d}\}$ and $\{x^{'}_1,..,x^{'}_{X_d}\}$ respectively. Then diversity is formalized as:
\begin{equation*}
    \text{DIV} = \frac{1}{X_d} \sum_{i=1}^{X_d} || x_i - x^{'}_{i} ||.
\end{equation*}
To evaluate MultiModality, a set of text descriptions with size $J_m$ is randomly sampled from all descriptions. Then two subsets of the same size $X_m$ are randomly sampled from all motions generated by $j\text{-th}$ text descriptions, with motion feature vectors $\{x_{j,1},..,x_{j,X_m}\}$ and $\{x^{'}_{j,1},..,x^{'}_{j,X_m}\}$ respectively. The multimodality is calculated as:

\begin{equation*}
    \text{MM} = \frac{1}{J_m \times X_m} \sum_{j=1}^{J_m} \sum_{i=1}^{X_m}|| x_{j,i}- x^{'}_{j, i} ||.
\end{equation*}

}

\textbf{Condition Matching}.
\jb{For the text-to-motion task, \cite{Guo_2022_CVPR_t2m} provides motion/text feature extractors to produce geometrically closed features for matched text-motion pairs, and vice versa. Under this feature space, 
motion-retrieval precision (R Precision) first mix generated motion with 31 mismatched motions and then calculates the text-motion top-1/2/3 matching accuracy, and Multi-modal Distance (MM Dist) that calculates the distance between generated motions and text.
For action-to-motion, for each dataset \cx{a pretrained recognition model~\cite{guo2020action2motion,petrovich21actor}} is used to calculate the average motion Accuracy (ACC) for action categories.
}

\textbf{Time Costs}. To evaluate the computing efficiency of diffusion models, especially the inference efficiency, we propose Average Inference Time per Sentence (AITS) measured in seconds. In our case, we calculate AITS ($cf.$ \cref{fig:comp:time}) on the test set of HumanML3D~\cite{Guo_2022_CVPR_humanml3d}, set the batch size to one, and ignore the time cost for model and dataset loading parts.

\end{document}